\documentclass{IEEE-Con-Sys-mag}
\pdfoutput=1

{\title{ Risk-Aware Robotics:\\ Tail Risk Measures in Planning, Control, and Verification}
\author{Prithvi Akella*, Anushri Dixit*, Mohamadreza Ahmadi, Lars Lindemann,  Margaret P. Chapman, George J. Pappas, Aaron D. Ames, and Joel W. Burdick\\}}

\affil{P. Akella (\href{mailto:prithvi.akella@gmail.com}{prithvi.akella@gmail.com}), A. D. Ames (\href{mailto:ames@.caltech.edu}{ames@caltech.edu}), and J. W. Burdick (\href{mailto:jwb@robotics.caltech.edu}{jwb@robotics.caltech.edu}) are with the Department of Mechanical and Civil Engineering, California Institute of Technology, Pasadena, CA, USA.\\
A. Dixit (\href{mailto:anushri.dixit@princeton.edu}{anushridixit@ucla.edu}) is with the Department of Mechanical and Aerospace Engineering, University of California, Los Angeles, USA.\\
M. Ahmadi (\href{mailto:reza.ahmadi@gatik.ai}{reza.ahmadi@gatik.ai}) is with Gatik AI, 161 E Evelyn Ave, Mountain View, CA 94041. \\
L. Lindemann (\href{mailto:llindema@usc.edu}{llindema@usc.edu} is with the Department of Computer Science, University of Southern California, Los Angeles, CA, USA. \\
M. P. Chapman (\href{mailto:mchapman@ece.utoronto.ca}{mchapman@ece.utoronto.ca}) is with the Edward S. Rogers Sr. Department of Electrical and Computer Engineering, University of Toronto, Toronto, ON, Canada.\\
G. J. Pappas (\href{mailto:pappasg@seas.upenn.edu}{pappasg@seas.upenn.edu}) is with the Department of Electrical and Systems Engineering, University of Pennsylvania,
Philadelphia, PA, USA. \\
\vspace{-0.8 in}
*Both authors contributed equally.}

\usepackage{amsthm}
\usepackage{amssymb}
\usepackage{amsmath}
\usepackage{mathrsfs}
\usepackage{hyperref}
\usepackage{makecell}
\usepackage{mathtools}
\usepackage{slashbox}
\usepackage{balance}

\usepackage[most]{tcolorbox}
\usepackage{graphicx}
\hypersetup{
    unicode=false,          % non-Latin characters in Acrobat’s bookmarks
    pdftoolbar=true,        % show Acrobat’s toolbar?
    pdfmenubar=true,        % show Acrobat’s menu?
    pdffitwindow=false,     % window fit to page when opened
    pdfstartview={FitH},    % fits the width of the page to the window
    pdfnewwindow=true,      % links in new PDF window
    colorlinks=true,       % false: boxed links; true: colored links
    linkcolor=blue,          % color of internal links (change box color with linkbordercolor)
    citecolor=blue,        % color of links to bibliography
    filecolor=magenta,      % color of file links
    urlcolor=red,           % color of external links
    breaklinks=true
}
\usepackage{cleveref}
\usepackage{xcolor}
\usepackage[sort, numbers, compress]{natbib}
\usepackage{here}
\usepackage{caption}
\usepackage{subcaption}

\usepackage{tikz}
\usetikzlibrary{shapes.geometric, arrows}
\tikzstyle{arrow} = [thick,->,>=stealth]
\tikzstyle{arrow_nohead} = [thick,-,>=stealth]
\tikzstyle{startstop} = [rectangle, rounded corners, minimum width=1cm, minimum height=2.5cm,text centered, text width=3cm, draw=black, fill=red!30]

\usepackage{tikz}
\usepackage{tikzscale}
\usepackage{pgfplots}
\usepackage{tikz}
\usepackage{xcolor}
\usepackage{graphicx}
\usetikzlibrary{automata, positioning, arrows}
\tikzset{
->, % makes the edges directed
>=latex,
node distance=3.5cm, % specifies the minimum distance between two nodes. Change if necessary.
every state/.style={thick, fill=blue!15}, % sets the properties for each ’state’ node
initial text=$ $, % sets the text that appears on the start arrow
}

\newcommand\equalsindist{\stackrel{\mathclap{\tiny\mbox{d}}}{=}}

\DeclareMathOperator*{\argmin}{argmin}
\DeclareMathOperator*{\suchthat}{~\mathrm{s.t.}~}

\DeclareMathOperator{\prob}{\mathbb{P}}

\DeclareMathOperator{\var}{VaR}
\DeclareMathOperator{\cvar}{CVaR}
\DeclareMathOperator{\evar}{EVaR}

\DeclareMathOperator{\uniform}{\mathrm{U}}
\DeclareMathOperator{\indicator}{\boldsymbol{1}}

\DeclareMathOperator*{\esssup}{ess\,sup}
\DeclareMathOperator{\trajectory}{\boldsymbol{x}}

\DeclareMathOperator{\classifier}{\boldsymbol{C}}
\DeclareMathOperator{\signalspace}{\mathcal{S}}
\DeclareMathOperator{\empiricaldist}{\hat{F}_N}
\DeclareMathOperator{\trajspace}{\mathcal{S}^{\mathcal{X}}}

\newcommand{\spacing}{\vspace{0.1 cm}}

\newcommand{\newidea}[1]{\noindent \underline{\textit{#1}}:}
\newcommand{\addendum}[1]{{\color{black} #1}}

\newtheorem{defin}{\bf Definition}
\newtheorem{theorem}{Theorem}

\newtheorem{remark}{Remark}

\newtheorem{problem}{Sidebar}
\newenvironment{open_prob}
  { \begin{tcolorbox}[
 colframe=yellow!70!white,
 colback=yellow!17!white,
 arc=8pt,
 breakable,
 left=1pt,right=1pt,top=1pt,bottom=1pt,
 boxrule=0.3pt,
 ]
\begin{problem}}
  {\end{problem}
 \end{tcolorbox}}

\begin{document}

\maketitle

\textbf{Why is the consideration of risk integral to robotics?} Consider the scenario of a rescue robot designed to navigate through a constrained environment to locate and rescue victims. Despite limited visibility and a ground that is potentially difficult to traverse, the robot is required to assess whether to execute a planned route or to seek an alternative route. Delays in finding an alternative route could negatively impact the victim's well-being while opting for the planned route could lead to the robot getting stuck. How should the robot gauge these competing risks? The issue of risk assessment is not new to roboticists and control engineers, and in practical applications, it is usually tackled using heuristics. For instance,  the robot constructs a map of the environment using computer vision. As this map is uncertain, one may attempt to decrease the robot's risk by tightening the area that can safely be traversed. This tightening is often based on past experimental data. However, this experimental data may not accurately represent real-world disaster scenarios. Additionally, an overly conservative tightening might lead the robot to waste time seeking alternate routes, while too little tightening could place the robot at risk of entrapment. Consequently, there is a need for a systematic approach to assessing the risks and rewards associated with different actions amid uncertainty. 

The need for a systematic approach to risk assessment has only increased in recent years due to the ubiquity of autonomous systems that alter our day-to-day experiences and their need for safety, e.g., for self-driving vehicles, mobile service robots, and bipedal robots. These systems are expected to function safely in unpredictable environments and interact seamlessly with humans, whose behavior is notably challenging to forecast. To reason about risk in such settings, the fields of systems science and control engineering have a long history and a rich literature on analyzing and designing systems under uncertainty. Existing methodologies can be broadly classified into the three paradigms of worst-case, risk-neutral, and risk-aware approaches as classified in~\cite{wang2022risk}. In the \emph{worst-case paradigm}, a system's ability to remain safe or perform satisfactorily is judged by examining its most severe safety violation or worst performance. This paradigm forms the basis of robust control~\cite{bertsekas1971minimax, zhou1998essentials, rawlings2017model} and robust safety analysis~\cite{chen2018hamilton}. For instance, if a system is supposed to track a planned trajectory, the largest discrepancy between the system’s realized trajectory and the planned trajectory can serve as a measure of the system’s performance. Contrarily, in the \emph{risk-neutral paradigm}, a system's capacity to stay safe or perform satisfactorily is evaluated on average or probabilistically.\footnote{Note that the probability of an undesirable event can be expressed as the expected value over an indicator function, making probabilistic reasoning conceptually similar to average reasoning.} This paradigm is often used in stochastic control and reinforcement learning \cite{bertsekas2004stochastic, sutton2018reinforcement, rawlings2017model} as well as in verification~\cite{abate2008probabilistic}. In the case of trajectory tracking, one would consider the average discrepancy between the system’s realized trajectory and the planned trajectory to gauge the system's performance when using a risk-neutral approach. If the system aims to reach a designated target while avoiding an unsafe region, the probability of the system accomplishing these goals can be used to assess the system's performance \cite{summers2010verification}. However, the worst-case paradigm may result in an overly conservative risk assessment and impractical solutions, while the risk-neutral paradigm can not account for harmful but less likely events. \addendum{We also note that the developments in the worst-case and the risk-neutral paradigms were often mathematically motivated but not necessarily practically driven as one was able to derive closed-form analytical solutions, e.g., as in linear robust control and linear quadratic control \cite{skogestad2005multivariable,zhou1998essentials}.}  The \emph{risk-aware paradigm}, on the other hand, lies conceptually in between and aims to evaluate a system's performance by giving attention to system outcomes that do not correspond to the worst case or the average.  Historically, the risk-aware paradigm has focused on the application of mean-variance approximations, e.g., for controller design~\cite{howard1972risk, jacobson1973optimal, whittle1981}, which is derived from the Markowitz model for evaluating the risk of financial portfolios~\cite{Markowitz1952}. 

A contemporary risk-aware approach that is adopted in this survey to mitigate rare and detrimental outcomes employs the use of \emph{tail risk measures}, a concept borrowed from financial literature \cite{rockafellar2000optimization}.  As such, this survey will introduce these measures and explain their relevance in the context of robotic systems.  To preface this explanation, however, we will first provide a deeper overview of the aforementioned paradigms, including an emphasis on the limitations of the \emph{worst-case} and \emph{risk-neutral} paradigms, which prompted the rise of \emph{risk-aware} study.

\addendum{\textbf{Benefits of the risk-aware paradigm}  Many systems are affected by uncertainties that can be well-modeled by random noise.  For example, uneven terrain can disturb a robot's planned trajectory~\cite{guzzi2019impact, karlsson2023risk}, wind can destabilize an aerial vehicle~\cite{akella2023learning}, and smoke can interfere with perception during an autonomous rescue mission~\cite{dixit2023step}.  However, designing operating rules to optimize a system's performance on average need not yield trustworthy or reliable performance in practice~\cite{ahmadi2020risk}.  For example, consider the disaster scenario presented in~\cite{dixit2023step}.  The robot must be able to assess risks in debris-laden zones while taking into account uncertainties derived from sensor and state estimation inaccuracies. If the robot operates under a worst-case scenario strategy, it may not find a feasible path through the rubble due to its aversion to any risk. On the other hand, a risk-neutral path-planning approach might expose the robot to unsafe, mission-jeopardizing situations. This example is among the many that underline the limitations of the risk-neutral and worst-case paradigms in the context of robotic applications. Another example is how humans may be risk-averse when collaborating with robots, requiring the robot to interpret the risk-averse human model for successful collaborations~\cite{kwon2020humanrisk}.  Case studies from our previous works will be presented later on to further illustrate these points, though we will start by introducing the tail risk measures that have helped foster the more recent, risk-aware paradigm.}

\begin{figure}
\centering{
    \resizebox{0.5\textwidth}{!}{
\begin{tikzpicture}
\tikzstyle{every node}=[font=\Large]
\pgfmathdeclarefunction{gauss}{2}{%
  \pgfmathparse{1000/(#2*sqrt(2*pi))*((x-.5-8)^2+.5)*exp(-((x-#1-6)^2)/(2*#2^2))}%
}
\pgfmathdeclarefunction{gauss2}{3}{%
\pgfmathparse{1000/(#2*sqrt(2*pi))*((#1-.5-8)^2+.5)*exp(-((#1-#1-6)^2)/(2*#2^2))}
}
\begin{axis}[
  no markers, domain=0:16, range=-2:8, samples=200,
  axis lines*=center, xlabel=$X$, ylabel=$P(X)$,
  every axis y label/.style={at=(current axis.above origin),anchor=south},
  every axis x label/.style={at=(current axis.right of origin),anchor=west},
  height=5cm, width=17cm,
  xtick={0}, ytick=\empty,
  enlargelimits=true, clip=false, axis on top,
  grid = major
  ]
   \addplot [fill=cyan!15, draw=none, domain=5:5.05] {gauss(1.5,2)} \closedcycle;
  \addplot [fill=cyan!20, draw=none, domain=7:15] {gauss(1.5,2)} \closedcycle;
    \addplot [fill=cyan!40, draw=none, domain=10:15] {gauss(1.5,2)} \closedcycle;
    \addplot [fill=cyan!65, draw=none, domain=13:15] {gauss(1.5,2)} \closedcycle;
  \addplot [very thick,cyan!50!black] {gauss(1.5,2)};
%   \addplot [very thick,cyan!50!black] {gauss(6.5,1)};
 
 \pgfmathsetmacro\valueA{gauss2(5,1.5,2)}
 \draw [gray] (axis cs:5,0) -- (axis cs:5,\valueA);
  \pgfmathsetmacro\valueB{gauss2(10,1.5,2)}
  \draw [gray] (axis cs:4.5,0) -- (axis cs:4.5,\valueB);
    \draw [gray] (axis cs:10,0) -- (axis cs:10,\valueB);
 
  \draw [gray] (axis cs:1,0)--(axis cs:5,0);
\node[below] at (axis cs:7.0, -0.1)  {$\mathrm{VaR}_{\beta}(X)$}; 
\node[below] at (axis cs:5, -0.1)  {$\mathbb{E}(X)$}; 
\node[below] at (axis cs:10, -0.1)  {$\mathrm{CVaR}_{\beta}(X)$}; 
\node[below] at (axis cs:13, -0.1)  {$\mathrm{EVaR}_{\beta}(X)$};
% \draw[] (0,1) node[below,xshift=4.5cm] {$\mathbb{E}(h)$} ;
\draw [yshift=2cm, latex-latex](axis cs:7,0) -- node [fill=white] {Probability~$\beta$} (axis cs:16,0);
\end{axis}
\end{tikzpicture}
}
\caption{\addendum{Visualization of common tail risk measures. The tail risk measures referenced in the article are shown above, applied to the random variable $X$ with distribution function $P$. $X$ could represent any cost random variable, for example - negative distance from an obstacle, distance from a goal, or energy used by the robot. Figure adapted from~\cite{dixit2022riskaverse}.}}
\label{fig:risk_fig}}
\end{figure}
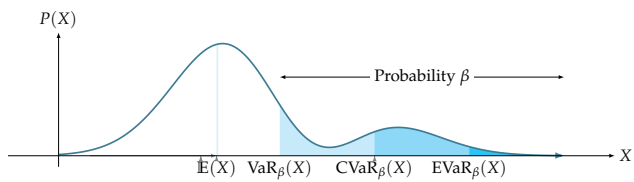

\textbf{Tail Risk Measures}  Succinctly, risk measures are functions over scalar-valued random variables designed to identify characteristics of interest of the random variable in question \cite{shapiro2009lectures}. By \emph{tail risk measures}, we imply that the risk measures of interest assess the upper right tail of the distribution of the random variable. Typically, the random variable indicates a cost so that tail risk measures capture the risk of incurring a high cost.   Figure~\ref{fig:risk_fig} depicts a few examples of such tail risk measures, namely, Value-at-Risk, Conditional-Value-at-Risk, and Entropic-Value-at-Risk. The Value-at-Risk at level $\beta \in (0,1)$ corresponds to the cutoff value for which a fraction $\beta$ of the outcomes of the random variable lies to the right of this cutoff value. A more formal description of these risk measures will follow in the Section ``Tail Risk Measures: Definitions and Notation.'' We focus on these measures though as they provide a systematic way of assessing the rare and unsafe (costly) outcomes that must be limited during planning, control, and verification. For example, consider a robot traversing through an environment with uncertain information about the obstacles' positions due to sensor noise. In this case, we could define a cost random variable by negating the minimum distance to all obstacles.  As such, more positive, \textit{c.f.} more costly outcomes, would correspond to more unsafe behavior as the system is not maintaining the required distance to the measured obstacle.  Then, as we have uncertain knowledge of the obstacle's location, we aim to minimize the tail risk incurred by this random cost in an effort to realize safe, risk-aware control actions.  This intuitive approach to risk-aware decision selection can be applied to several facets of an autonomy stack, \textit{i.e.} planning, control, and verification, as has recently been done in both the controls and robotics communities.

\textbf{Organization} As such, this survey serves as an introduction to risk-aware planning, control, and verification in robotics, employing tail risk measures - an emerging field in the literature. Our focus areas include:
\begin{itemize}
    \item A summary of risk measure theory with an emphasis on tail risk measures in the section "Tail Risk Measures: Definitions and Notation". We highlight how these measures offer a consistent and intuitive means to adjust the system's risk aversion levels.
    \item A discussion on the key principles of risk-aware planning and control, an introduction of the algorithms, and multiple presentations of real-world case studies such as planning and control in subterranean environments. These are covered in the sections "Risk-Aware Planning" and "Risk-Aware Control".
    \item A review of temporal logics as a mathematical formalism for articulating complex robotic system specifications in the "Verification" section. Additionally, we explore why it is important to consider the tail risk of system trajectories evaluated against these specifications.
    \item An introduction of a tail risk method for safety-critical controller verification in the "Verification" section. We demonstrate this risk-aware verification approach's ability to effectively identify potential mission issues while validating the probabilistic verification statements made within the described framework.
    \item We end the survey with open problems and future research directions.
    \end{itemize}

\textbf{Related Work} Despite the crucial need for systematic risk evaluation in robotics applications, recent surveys do not emphasize risk-aware planning, control, and verification \cite{schwarting2018planning, karpas2020automated, brunke2022safe}. For instance, Schwarting, Alonso-Mora, and Rus examine planning and decision-making methods for autonomous driving \cite{schwarting2018planning}, Karpas and Magazzeni discuss strategies that enable robots to automatically combine smaller tasks to achieve broader goals \cite{karpas2020automated}, and Brunke et al. outline the interplay between control theory and reinforcement learning for safety in robotic applications \cite{brunke2022safe}. However, these works do not focus on risk measures. Moreover, Hobbs et al. provide a comprehensive introduction to non-stochastic run-time assurance systems, such as a system that overrides an existing controller when an extreme hazard is detected, without an emphasis on risk measures \cite{Hobbs2022}. Two closely related works to our survey are Majumdar and Pavone's \cite{majumdar2020should} and Wang and Chapman's \cite{wang2022risk}. Majumdar and Pavone propose axioms for risk measures suitable for robots and provide intuitive explanations for these axioms. However, their work does not take the form of a survey and present algorithms for risk-aware planning, control, and verification \cite{majumdar2020should}. On the other hand, Wang and Chapman overview historical and modern research about risk-aware autonomous systems, but they do not focus on robotics applications, temporal logics, or certain tail risk measures such as entropic value-at-risk and total variation distance-based risk measure \cite{wang2022risk}. Our survey draws inspiration from Majumdar and Pavone \cite{majumdar2020should} and Wang and Chapman \cite{wang2022risk} and aims to elucidate the concept of tail risk measures for the control systems community and to showcase their utility for planning, control, and verification of robotic systems. We provide the much-needed emphasis on these risk measures in robotics, helping to ensure that the development and application of autonomous systems remain safe, effective, and mindful of potential risks. As we move on in this survey, we will provide more pointers to relevant literature.

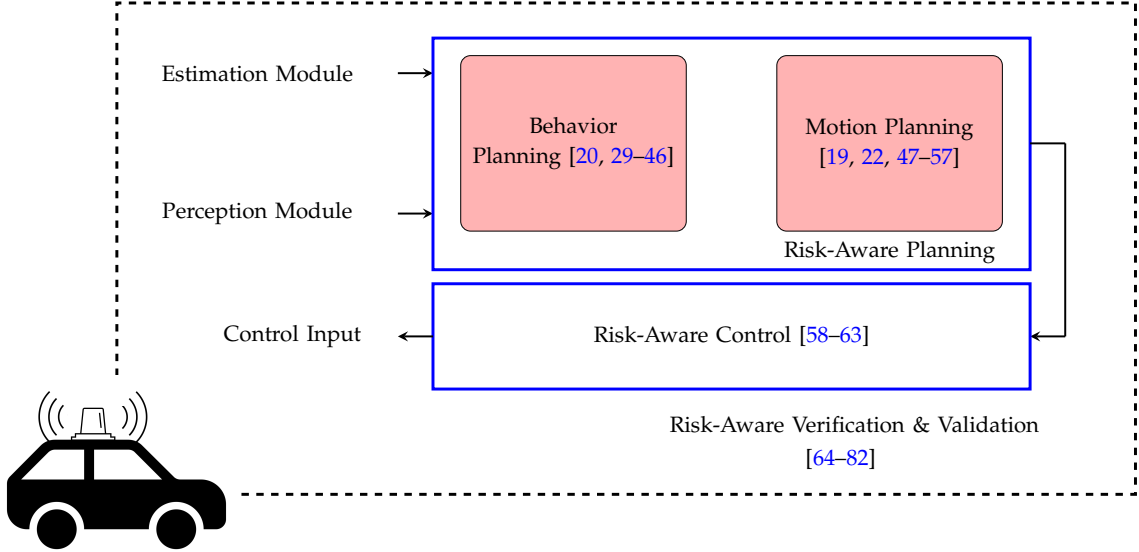
\begin{figure*}[ht]
\centering{
    \resizebox{0.9\textwidth}{!}{
\begin{tikzpicture}
\draw[black, very thick, dashed] (0,0) rectangle (14.5,7);
\node at (2,6) {Estimation Module};
\draw [arrow] (4,6) -- (4.5,6);
\node at (2,4) {Perception Module};
\draw [arrow] (4,4) -- (4.5,4);
\node[inner sep=0pt] (russell) at (0,0)
    {\includegraphics[width=.2\textwidth]{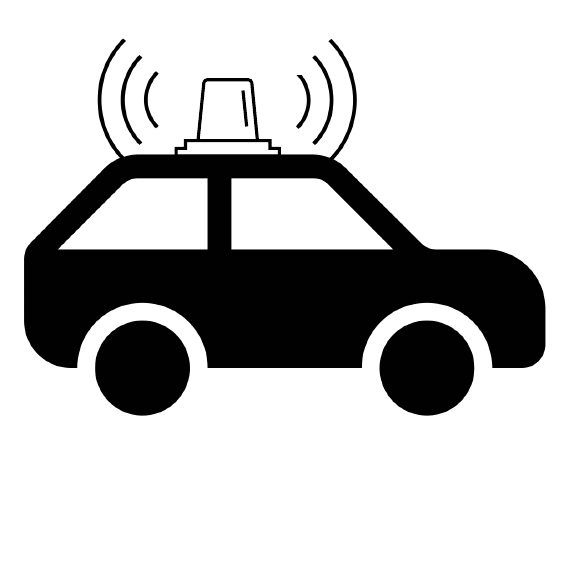}};
\draw[blue, very thick] (4.5,3.2) rectangle (13,6.5);
% % \node at (4.5,5) (start) [startstop] {Uncertainty processing (cite)};
% \draw [arrow] (5.75,5) -- (6.25,5);
\node at (6.5,5) (start) [startstop] {Behavior Planning~\cite{chow2017risk,prashanth2022risk,bauerle2014more,prashanth2014policy,chow2014algorithms,tamar2016sequential,tamar2015policy,haskell2015convex,feyzabadi2014risk,pereira2013risk,lam2023risk,hau2023dynamic,ahmadi2021constrained,carpin2016risk,gavriel2012risk,ahmadi2021risk,fan2018risk,ahmadi2020risk,ahmadi2023risk} };
\draw [arrow] (9.75,5) -- (10.25,5);
\node at (11,5) (start) [startstop] {Motion Planning

\cite{singh2018framework,dixit2022tvd,parys2016DRCC,chen2022interactive,Sopasakis2019,coulson2019data, coulson2021distributionally,zolanvari2022risklmpc,hakobyan2019risk,dixit2020risksensitive,dixit2022riskaverse,dixit2023step, fan2021step}};
\node at (11,3.45) {Risk-Aware Planning};
\draw [arrow_nohead] (13,5) -- (13.5,5);
\draw [arrow_nohead] (13.5,5) -- (13.5,2.25);
\draw [arrow] (13.5,2.25) -- (13,2.25);
\draw[blue, very thick] (4.5,1.5) rectangle (13,3);
\node at (2.5,2.25) {Control Input};
% \node at (7, 2) {(cite papers here)};
\node at (8.75,2.25) {Risk-Aware Control~\cite{SamuelsonInsoon2018, chapman2019risk, chapman2021risk, chapman2022optimizing, weifausschapmanconf2022,singletary2022safe}};
\draw [arrow] (4.5,2.25) -- (4,2.25);
\node at (10.5,1) {Risk-Aware Verification \& Validation};
\node at (10.25,0.5){~\cite{akella2022scenario,lindemann2021stl,cubuktepe2018verification,leung2022semi,meggendorfer2021verification,samuelson2018safety,hashemi2022risk,quagliarella2017robust,lindemann2022risk,lindemann2022temporal,akella2022sample,rigter2021risk,godbout2021carl,hong2011monte,kim2015conditional,kim2022efficient,dosovitskiy2017carla,akella2022barrier,akella2022safety}};
\end{tikzpicture}}}
\caption{An overview of a typical (risk-aware) planning and verification pipeline in an autonomy stack. }
\end{figure*}

\section{Tail Risk Measures: Definitions and Notation}\label{sec:risk_measures}
\addendum{
\textbf{A Pedagogical Note on the use of Measure Theory}  A formal definition of tail risk measures such as Definition~\ref{def:tail_risk_measures} to follow, requires some concepts from measure theory, namely distributional equivalence, \textit{i.e.} $\equalsindist$.  As such, we begin this section in a more measure-theoretic context, introducing probability spaces first and defining random variables over these spaces.  For readers interested in a deeper study into probability and measure theory, please reference~\cite{ash2000probability}.  For those unfamiliar, however, it is sufficient to consider the to-be-mentioned scalar random variables $X$ as functions assigning real numbers to the random outcomes of interest, \textit{e.g.} the canonical example of assigning the numbers $1-6$ to the corresponding rolls of a six-sided die.  Likewise, the distribution $P$ of this random variable $X$ denotes the probability of each sampleable event.  Continuing with the six-sided die example and assuming a fair die, the probability of realizing any specific roll - the probability of sampling any number $1-6$ under the random variable $X$ - would all be $\frac{1}{6}$.  With this intuition, we will begin with a formal definition of tail risk measures.}

Consider a probability space $(\Omega, \mathcal{F}, {P})$, where $\Omega$, $\mathcal{F}$, and ${P}$ are the sample space, the $\sigma$-algebra over $\Omega$, and the probability measure over $\mathcal{F}$, respectively. A random variable $X: \Omega\xrightarrow[]{}\mathbb{R}$ denotes the cost of each outcome, and $\mathcal{X}$ is the set of all such random variables defined on $\Omega$. For any random variable $X \in \mathcal{X}$, $F_X(x)$ refers to the cumulative distribution function with inverse $F_X^{-1}(\beta) = \inf\{x\in\mathbb{R}| F_X(x) \geq 1-\beta \}$. For any two random variables $X,X' \in \mathcal{X}$, the expression $X \equalsindist X'$ denotes that the random variables $X,X'$ have the same distribution under ${P}$. Similarly, we use $X\le X'$ as a shorthand notation to indicate that $X(\omega)\le X'(\omega)$ \addendum{for almost all} $\omega\in\Omega$.  Finally, $\uniform$ denotes the uniform random variable between $[0,1]$.

A risk measure $\rho$ is a function that maps a cost random variable to a real number, \textit{i.e.} $\rho:\mathcal{X}\xrightarrow[]{}\mathbb{R}$. Informally, tail risk measures refer to the behavior of the cost $X$ in the tail of its distribution.  Mathematically, consider the random variable $X \in\mathcal{X}$ and the risk-level $\beta \in (0,1)$. As seen in~\cite{fangda2021tailrisk}, we define $X_\beta$ to be the tail risk of $X$ such that,
\begin{align}
    X_\beta = F_X^{-1}(1-\beta + \beta \uniform).
\end{align}
In other words, the distribution of the random variable $X_\beta$ is the distribution of $X$ in its $\beta$-quantile (or tail) normalized to sum to $1$. It is also clear that as $\beta\xrightarrow[]{}0$, $X_\beta\xrightarrow[]{}\esssup (X)$. We are now ready to formally define tail risk measures.
\begin{defin}[Tail Risk Measures~\cite{fangda2021tailrisk}]
    \label{def:tail_risk_measures}
    For $\beta \in (0, 1)$, a risk measure $\rho$ is a $\beta$-tail risk measure (or simply \textit{tail risk measure}) if $\rho(X) = \rho(X')$ for all $X,X'\in\mathcal{X} $ satisfying $X_\beta \equalsindist X'_\beta$.  
\end{defin}
\noindent The aforementioned definition is a formal description of tail risk measures, three of which have featured more prominently in recent literature --- Value-at-Risk, Conditional-Value-at-Risk, and Entropic-Value-at-Risk.  Their definitions will follow.
\subsection{Value-at-Risk}
Chance constraints can be reformulated by a commonly used risk measure called the \textit{Value-at-Risk} (VaR). For a given confidence level $\beta \in (0,1)$, $\mathrm{VaR}_{\beta}$ denotes the $\beta$-quantile value of the cost variable $X$ and is defined as, 
\begin{align*}
    \text{VaR}_{\beta}(X) \coloneqq \inf \{ z \in\mathbb{R}\,|\,F_X(z) \geq 1-\beta\}.
\end{align*}
Therefore, $\text{VaR}_\beta(X) = F_X^{-1}(\beta)$. It follows that $
    \text{VaR}_{\beta}(X)\leq 0 \implies P(X \leq 0)\geq 1-\beta.$
    % VaR is nonconvex for general distributions and is not a coherent risk measure. 

\subsection{Conditional Value-at-Risk}

The {\em Conditional Value-at-Risk}, $\mathrm{CVaR}_{\beta}$, measures the expected loss in the ${\beta}$-tail of the random variable $X$.  Formally, for some $\beta \in (0,1]$ $\mathrm{CVaR}_{\beta}$ is defined as follows per~\cite{rockafellar2000optimization}:

\begin{equation}
\begin{aligned}
    \mathrm{CVaR}_{\beta}(X)\coloneqq&\inf_{z \in \mathbb{R}}\mathbb{E}\Bigg[z + \frac{(X-z)^{+}}{\beta}\Bigg]
    % \\ =& \inf_{z \in \mathbb{R}, s \in \mathbb{R}^J} \sum_{j=1}^{J}p(j) \bigg(z +\frac{s(j)}{\beta} \bigg) \\
    % & \quad\text{s.t.} \quad s(j) \geq 0, \, s(j) + z \geq X^j,%\\
    % =& \sup_{Q\in\mathcal{Q}}\mathbb{E}_Q(C)
\end{aligned}
\end{equation} 
where we use the notation $(X-z)^+=\max(0,X-z)$. A value of $\beta \simeq 1$ corresponds to a risk-neutral case. A value of $\beta \to 0$ is rather a risk-averse case\footnote{Another, more intuitive, way to think about the widely used $CVaR$ metric is that it is the expectation of the random variable $X$ conditioned on  $X\ge VaR_{\beta}(X)$, i.e., $CVaR_{\beta}(X) = \mathbb{E}[X|X\ge VaR_{\beta}(X)]$. For example, the $5 \%$ CVaR risk of a portfolio is equivalent to the expected (mean) return on a portfolio in the worst $5\%$ of scenarios over a specified time horizon (in this definition we assume that $X$ is larger for worse returns).}.  Importantly, CVaR is a coherent risk measure~\cite{artzner1999coherent} (see the text below),  prompting its widespread use in the recent literature.  Furthermore, CVaR is a loose upper bound on VaR, \textit{i.e.},
\begin{align}
    {\text{VaR}_{\beta}(X) \leq \text{CVaR}_{\beta}(X) \leq 0 \implies P(X\leq 0)\geq 1-\beta.}
\end{align}

\subsection{Entropic Value-at-Risk}
The {\em Entropic Value-at-Risk}, EVaR$_{\beta}$, derived using the Chernoff inequality for the random variable in question, is the tightest upper bound for VaR and CVaR.  It was shown in~\cite{ahmadi2017analytical} that $\mathrm{EVaR}_{\beta}$ and $\mathrm{CVaR}_{\beta}$ are equal \addendum{when  $\mathrm{VaR}_{t}(X) = -\infty$ for all $\beta < t \leq 1$.} For some $\beta \in (0,1]$, 
\begin{equation} \label{eq:evar}
   \mathrm{EVaR}_{\beta}(X)\coloneqq \inf_{z > 0 }\Bigg[z^{-1}\ln\frac{\mathbb{E}[e^{Xz}]}{\beta}\Bigg].
    % = \sup_{Q\in\mathfrak{D}}\mathbb{E}_Q(X).
\end{equation}
Similar to $\mathrm{CVaR}_{\beta}$, for $\mathrm{EVaR}_{\beta}$, the limit $\beta \to 1$ corresponds to a risk-neutral case; whereas, $\beta \to 0$ corresponds to a risk-averse case. In fact, it was demonstrated in~\cite[Proposition 3.2]{ahmadi2012entropic} that $\lim_{{\beta}\to 0} \mathrm{EVaR}_{{\beta}}(X) = \esssup(X)$, \addendum{where $\esssup(X)$ is the essential supremum of the random variable $X$}. Finally, EVaR is also a coherent risk measure and is an upper bound for both VaR and CVaR:
\begin{align*}
    \text{VaR}_{\beta}(X) &\leq \text{CVaR}_{\beta}(X) \leq \text{EVaR}_\beta(X) \leq 0, \\ & \implies P(X\leq 0)\geq 1-\beta.
\end{align*}
\addendum{EVaR gets its name from its dual representation as a risk measure that provides distributional robustness using relative entropy. }
\subsection{Coherent Risk Measures}
\addendum{Coherent risk measures are a prominent class of risk measures that are well-regarded for their robust mathematical properties, their inherent intuitiveness for risk analysis, and their use in robotics \cite{majumdar2020should}.} Introduced by Artzner et al.~\cite{artzner1999coherent} in the context of financial risk management, coherent risk measures satisfy four axiomatic properties: monotonicity, subadditivity, positive homogeneity, and translational invariance. Monotonicity indicates that adding a less risky outcome to a portfolio should not increase its risk. Subadditivity implies that diversifying a risk portfolio should not increase its overall risk. Positive homogeneity signifies that scaling all outcomes in a portfolio should proportionally scale its risk. Translational invariance denotes that adding a risk-free asset to a portfolio should decrease its risk correspondingly. Coherent risk measures offer a rich theoretical foundation to quantify and manage risk systematically. They enable us to capture extreme, but rare, high-consequence events and provide a mechanism to evaluate and compare different risk scenarios, making them particularly valuable for risk-aware planning, control, and verification in robotics. We are now ready to describe coherent risk measures. 

\begin{defin}
[Coherent Risk Measure]\label{defi:coherent}{
We call the risk measure $\rho: \mathcal{X}\to \mathbb{R}$, a \emph{coherent risk measure}, if it satisfies the following conditions
\begin{itemize}
    \item \textbf{Subadditivity:} $\rho( X + X') \le \rho(X)+\rho(X')$, for all $X,X' \in \mathcal{X}$;
    \item \textbf{Monotonicity:} If $X\le X'$ then $\rho(X) \le \rho(X')$ for all $X,X' \in \mathcal{X}$;
    \item \textbf{Translational Invariance:} $\rho(X+c)=\rho(X) + c$ for all $X \in \mathcal{X}$ and $c  \in \mathbb{R}$;
    \item \textbf{Positive Homogeneity:} $\rho(\beta X)= \beta \rho(X)$ for all $X \in \mathcal{X}$ and $\beta \ge 0$.
\end{itemize}
}
\end{defin}

Note that the properties of subadditivity and positive homogeneity together imply that coherent risk measures are also convex.
CVaR and EVaR have been recognized as coherent risk measures. Conversely, VaR does not qualify as a coherent risk measure, as it does not satisfy the sub-additivity property. This fact limits its utility in situations where a joint assessment of risks is required, which is often the case in risk-aware robotic planning, control, and verification.  While VaR is not a coherent risk measure, each of the measures defined above has its place in risk evaluation, depending on the specific requirements and constraints of the task at hand.

\addendum{Coherent risk measures can also be written as the worst-case expectation over a convex and closed set of probability mass (or density) functions. This distributionally-robust representation of coherent risk measures is defined below. 
\begin{defin}[Representation Theorem]{Every coherent risk measure can be represented in its dual form as, 
\begin{align*}
    \rho(X) := \sup_{Q \in \mathcal{Q}}E_Q(X),
\end{align*}
where the ambiguity set (or risk envelope), $\mathcal{Q}$, is convex and closed, and probability density function $Q(X)$ is absolutely continuous with respect to the probability density function $P(X)$; \textit{i.e.,} $P(X)=0\ \implies \ Q(X)=0$.}
\end{defin}

\begin{remark}
     In this review, we limit our focus to tail risk measures for robotics but we note that distributionally-robust optimization is another widely used measure of risk~\cite{rahimian2019distributionally, shapiro2021tutorial, safaoui2021riskaverserrtplanningnonlinear, RENGANATHAN2023103812, renganathan2020towards}. Some tail risk measures, like CVaR and EVaR, may have a distributionally-robust interpretation as well. Some methods that consider distributionally-robust formulations of tail risk measures are also discussed in this review. 
\end{remark}
}
Coherent risk measures provide a snapshot of potential perils based on current conditions. These measures fall short when applied to dynamic systems, such as those common in robotics, where risks and conditions vary with time.  {\em Dynamic coherent risk measures} extend beyond the static approach, taking into account the time-varying nature of risk. They are particularly adept at characterizing the evolving risk landscape in dynamic environments. Dynamic risk measures continuously monitor and update risk assessments in response to changes in the system and its environment. This capability to adapt and provide a comprehensive understanding of risk in a fluctuating context aligns well with the realities of robotic operations in unstructured environments. 

\addendum{To define these measures, we will first index a sequence of random variables $X_t,~t=0,\ldots, N$, where $N \in \mathbb{N}_{\ge 0} \cup \{\infty\}$.  Note that prior, each random variable $X$ was defined over a corresponding probability space, and $\mathcal{X}$ denoted the set of all random variables defined over that space $(\Omega,\mathcal{F},{P})$. However, in the dynamic context, each sequenced random variable $X_t$ belongs to (perhaps) different spaces of random variables $\mathcal{X}_t$.  As such, denote $\mathcal{X}_{t:N}=\mathcal{X}_t\times \cdots \times \mathcal{X}_N$ as the space in which the corresponding sequence $X_{t:N}$ lives and $\mathcal{X}=\mathcal{X}_0\times \mathcal{X}_1 \times \cdots$.  Finally, we assume that the sequence $\boldsymbol{X} \in \mathcal{X}$ is almost surely bounded (with exceptions having probability zero), \textit{i.e.}, $\max_t \esssup~| X_t(\omega) | < \infty.$}

To describe how one can evaluate the risk of sub-sequence $X_t,\ldots, X_N$ from the perspective of stage $t$, we require the following definitions.
\vspace{0.2cm}
\begin{defin}
    [Conditional Risk Measure]{
A mapping $\rho_{t:N}: \mathcal{X}_{t:N} \to \mathcal{X}_{t}$, where $0\le t\le N$, is called a \emph{conditional risk measure}, if it has the following monoticity property:
\begin{equation*}
    \rho_{t:N}(\boldsymbol{X}) \le   \rho_{t:N}(\boldsymbol{X}'), \quad \forall \boldsymbol{X}, \forall \boldsymbol{X}' \in \mathcal{X}_{t:N}~\text{such that}~\boldsymbol{X} \preceq \boldsymbol{X}'.
\end{equation*}
}
\end{defin}
\vspace{0.2cm}
\begin{defin}[Dynamic Risk Measure]
{A \emph{dynamic risk measure} is a sequence of conditional risk measures $\rho_{t:N}:\mathcal{X}_{t:N}\to \mathcal{X}_{t}$, $t=0,\ldots,N$.}
\end{defin}
\vspace{0.2cm}
A key attribute of dynamic risk measures is their temporal consistency~\cite[Definition 3]{ruszczynski2010risk}: if two scenarios $X$ and $X'$ are identical for a time interval $[\tau, \theta]$, and if $X$ is evaluated to be as favorable as $X'$ at some future time point  $\theta$, then it stands to reason that $X$ should not be viewed as more risky than $X'$ at the earlier time point $\tau$. This principle ensures that the risk assessment remains coherent and consistent across different points in time. The definition of a dynamic coherent risk measure~\cite[p. 298]{shapiro2014lectures} then follows from its static counterpart.

\begin{defin}[Dynamic Coherent Risk Measure]\label{defi:coherent}{
We call the one-step conditional risk measures $\rho_t: \mathcal{X}_{t+1}\to \mathcal{X}_t$, $t=1,\ldots,N-1$ a \emph{coherent risk measure} if it satisfies the following conditions
\begin{itemize}
    \item \textbf{Convexity:} $\rho_t( X + (1-\lambda)X') \le \lambda \rho_t(X)+(1-\lambda)\rho_t(X')$, for all $\lambda \in (0,1)$ and all $X,X' \in \mathcal{X}_{t+1}$;
    \item \textbf{Monotonicity:} If $X\le X'$ then $\rho_t(X) \le \rho_t(X')$ for all $X,X' \in \mathcal{X}_{t+1}$;
    \item \textbf{Translational Invariance:} $\rho_t(X+X')=X+\rho_t(X')$ for all $X \in \mathcal{X}_t$ and $X' \in \mathcal{X}_{t+1}$;
    \item \textbf{Positive Homogeneity:} $\rho_t(\beta X)= \beta \rho_t(X)$ for all $X \in \mathcal{X}_{t+1}$ and $\beta \ge 0$.
\end{itemize}
}
\end{defin}

\section{An Overview of Tail Risk Measures in Robotics}
\addendum{The definitions of the prior section now permit us to provide a general overview of how tail risk measures are used to generate risk-aware behavior or validate systems in a risk-aware setting.  For such an overview, let us define $\rho$ to be a risk measure that} maps a random variable $X$ to a real number that indicates the risk associated with $X$. \addendum{For applications in robotics, this} random variable $X$ denotes any random cost associated with (perhaps) stochastically evolving system trajectories. To be more specific, consider a (perhaps) nonlinear discrete-time control system  at time $t$ with state $\boldsymbol{x}(t) \in \mathcal{X}\subseteq \mathbb{R}^n$, input $\boldsymbol{u}(t) \in \mathcal{U}$, and system disturbances $\boldsymbol{d}(t) \sim \xi(\boldsymbol{x}(t),\boldsymbol{u}(t),t)$ where $\xi(\boldsymbol{x}(t),\boldsymbol{u}(t),t)$ is a (perhaps) state, input, and time-dependent probability distribution over $\mathbb{R}^n$:
\begin{equation}
    \label{eq:gen_sys_update}
    \boldsymbol{x}(t+1) = f(\boldsymbol{x}(t),\boldsymbol{u}(t),\boldsymbol{d}(t)).
\end{equation}
Given a feedback controller $U:\mathcal{X} \to \mathcal{U}$, \addendum{which prescribes $\boldsymbol{u}(t) = U(\boldsymbol{x}(t))$,} we can construct a closed-loop, stochastically evolving dynamical system.
\addendum{To be clear, we first define $\signalspace^{\mathcal{X}}$ as the set of all signals mapping time to the state space $\mathcal{X}$, \textit{e.g.} $\signalspace^{\mathcal{X}} = \{s : \mathbb{Z} \to \mathcal{X}\}$.  As such, provided an initial condition $\boldsymbol{x}_0 \in \mathcal{X}$, this feedback controller $U$, and a randomly sampled disturbance sequence $\{\boldsymbol{d}(0),\boldsymbol{d}(1),\dots\}$, the corresponding state trajectory $\trajectory = \{\boldsymbol{x}(0)=\boldsymbol{x}_0, \boldsymbol{x}(1), \dots\}$ following the update in~\eqref{eq:gen_sys_update} is a signal that lives in $\signalspace^{\mathcal{X}}$, \textit{i.e.} $\trajectory \in \signalspace^{\mathcal{X}}$.  Notably, this trajectory $\trajectory$ is also a randomly sampled trajectory conditioned on the initial state $x_0$ due to the randomly sampled disturbance sequence.  To be more specific in a tail-risk context then, we define $\Sigma(\boldsymbol{x}_0)$ as the random variable conditioned on the initial condition $\boldsymbol{x}_0$ whose samples are these random trajectories $\trajectory$ living in $\signalspace^{\mathcal{X}}$, \textit{i.e.},
}
\begin{equation}
\label{eq:trajectory_sample}
\mathrm{a~sample~of~}\Sigma(\boldsymbol{x}_0)~\mathrm{is~} \trajectory \triangleq \{ \boldsymbol{x}(0)=\boldsymbol{x}_0, \boldsymbol{x}(1), \dots\},  
\end{equation}
To analyze the tail risk \addendum{of this trajectory random variable $\Sigma(x_0)$}, we require a mapping from trajectory samples $\trajectory$ to the real line:
   \[ C:\trajspace \to\mathbb{R}.\]
The function $C$ may denote the inverse distance of a robot's trajectory to an obstacle and thereby encode the system's robustness to a collision.  These cost functions can be constructed from principled approaches when the system's desired behavior is expressed as a temporal logic specification. \addendum{Applying this cost $C$ to the random variable corresponding to system trajectories $\Sigma(x_0)$ defines the scalar random variable $X$ over which we can apply our tail risk measure $\rho$, \textit{i.e.} $X = C(\Sigma(x_0))$.} For robotic planning, control, and verification, one can then minimize the risk of this cost, $\rho(X)$, or consider using risk as a constraint--$\rho(X)\le r$ for a risk threshold $r$.

\addendum{In what follows, we present the two main sections ``Risk-Aware Planning and Control'' and ``Risk-Aware Verification and Validation'', in this order. We chose this specific organization as planning and control synthesis is usually carried out by making simplifying assumptions on the system and by considering low-fidelity models. The design using low-fidelity models is motivated by the complexity of the problem and computational considerations. The designed planners and controllers are hence not guaranteed to be correct for high-fidelity models or real robotic systems, emphasizing the need for system verification and validation techniques that apply in realistic settings.}

\section{Risk-Aware Planning and Control}

The robotic behavior, motion planning, and control problems focus on designing algorithms that allow a robot to interact intelligently and safely with its surroundings. This complex process involves deciding on possible actions, constructing trajectories that a robot can take to achieve specific high-level goals, e.g., safely navigating through unstructured environments, and controlling the instantaneous robot motions to track the trajectory. Behavior planning concerns the higher-level decision-making needed to achieve higher-level robot objectives, while motion planning plans the details of robot movement, determining how a robot should move from one location to another while avoiding obstacles, and factoring in the robot's kinematics and dynamics. The control level manages the details of motion execution by continually computing the system control inputs that minimize tracking error while accounting for uncertainty and unexpected events. 

The process of designing robot plans and controls should critically consider potential risks and their overall system implications. These risks include physical risks to humans close to a robot as well as risks to the robot itself due to its unpredictable environment and imperfect robot sensing and perception systems. Notions of risk are particularly important in high-stakes robot tasks, such as search and rescue, or exploration of hazardous sites.  To properly manage these risks, this paper advocates for the use of tail risk measures. Importantly, tail risk measures focus on more extreme but rarer events, and thus provide a systematic and principled approach to quantifying and managing risk in robotics. Integrating tail risk measures into robotic planning and control modules can lead to more robust and safer robot operation, effectively balancing performance and safety under uncertainty.

\vskip 0.07 true in
\newidea{Historical Remark on Exponential Utility} (This remark is adapted from \cite{chapman2021classical}.)
Risk-aware control has been in development for at least fifty years. The earliest contributions concern the exponential utility measure (i.e., entropic risk measure):
\begin{equation}
    \rho_{\text{EU},\theta}(X) \coloneqq \textstyle{\frac{-2}{\theta}} \log \mathbb{E}\left[\exp(\textstyle{\frac{-\theta}{2}}X)\right],
\end{equation}
where $\theta \neq 0$ is a risk parameter and $X$ is a nonnegative random variable, which can be interpreted as a mean-variance approximation \cite{whittle1981risk}. When $\theta<0$, the robotic system is risk averse, while the robot will exhibit more risk tolerant behavior when $\theta>0$.   When $\theta=0$, the system is risk neutral. Notably, the exponential utility is \emph{not} a tail risk measure.
%, though it does represent one of the earliest notions of risk-aware control.  
To our knowledge, the first paper in the area of risk-aware control was a 1972 study about finite-state Markov decision processes by Howard and Matheson, in which performance was assessed by an exponential utility criterion~\cite{howard1972risk}.   The authors took inspiration from game theory \cite{luceraiffa}. One year later, Jacobson investigated the exponential utility criterion in the classical linear-quadratic setting (linear dynamics, quadratic costs, Euclidean spaces, and additive Gaussian noise) \cite{jacobson1973optimal}. Whittle developed key contributions regarding risk-aware control in the linear-quadratic setting using the exponential utility measure, including showing close analogies to optimal control and state estimation results in the risk-neutral case \cite{whittle1981risk, whittle1990risk}. While exponential utility continues to be investigated (e.g., see \cite{saldi2020approximate, blancas2020discounted, chapman2021classical}), recent attention focuses on different types of risk-aware performance and safety criteria, motivating this survey on tail risk. 

\subsection{Risk-Aware Behavior Planning}\label{subsection:high_level_planning}
\addendum{Behavior planning for robotics has been studied in many mathematical frameworks, though the most commonly utilized framework has been to reframe planning as a (partially observed) Markov Decision Problem (PO)MDP, which we will discuss in this section.} Specifically, an MDP is a triple\footnote{\addendum{Here, we note that MDPs are typically defined with an associated objective function, and we will define one for risk-aware behavior planning in a section to follow.}},
\begin{equation}
    \mathcal{M} = (\addendum{\mathcal{X}, \mathcal{U}}, T),
\end{equation}
where \addendum{$\mathcal{X} = \{x_{1},\dots,x_{|\mathcal{X}|}\}$} represents the states of the autonomous agent(s) and the world model, \addendum{$\mathcal{U}= \{u_{1},\dots,u_{|\mathcal{U}|}\}$} represents the actions available to the agent, and $T(\addendum{x_{j} |x_{i}, u})$ defines the likelihood of transitioning to state \addendum{$x_j$} from state \addendum{$x_i$} when taking action \addendum{$u \in \mathcal{U}$}. Furthermore, the transition function must be such that:
\begin{equation}
    \sum_{\addendum{x  \in \mathcal{X}} } T(\addendum{x |x_{i} , u}) = 1, \forall~\addendum{x_i  \in \mathcal{X} ,\forall~u \in \mathcal{U}},
\end{equation}
and for a \textit{finite} Markov Decision Process, the state and action spaces must also be finite.  Finally, a policy $\pi$ maps states to actions, \textit{i.e.} $\pi: \addendum{\mathcal{X} \to \mathcal{U}}$.

\addendum{As such, risk-aware behavior planning aims to construct a policy $\pi$ that minimizes a risk-measure evaluation over either each state or over some time-horizon of states.  Specifically, consider a cost $c$ associated with transitioning from an initial state $x_0$ via taking an action $u$ to a final state $x_f$, \textit{i.e.} $c: \mathcal{X} \times \mathcal{U} \times \mathcal{X} \to \mathbb{R}$ where $c(x_0,u,x_f)$ denotes the cost of the action-specific transition.  As state evolution for an MDP is stochastic, the exact cost of an action is unknown \textit{apriori} as the final state $x_f$ will be a sample from a random variable $X_f(x_0,u)$ conditioned on the initial state and action choice with distribution provided by the transition function $T$.  As a result, the corresponding cost of an initial state and action pair is a scalar random variable corresponding to the costs of the random final state samples, \textit{i.e.} $C(x_0,u) = c(x_0,u,X_f(x_0,u))$.  As such, the one-step optimal policy is to choose actions per state that minimize the tail-risk $\rho$ associated with this cost random variable $C(x_0,u)$, \textit{i.e.}
\begin{equation}
    \label{eq:risk_aware_policy_MDP}
    \pi^*(x) = \arg\min_{u \in \mathcal{U}} \rho(C(x, u)).
\end{equation}
We could similarly define policies by considering not just the one-step risk, but the risk over a horizon.  To do so, we note that for any horizon length $T \in \{1,2,\dots\} \triangleq = \mathbb{Z}_+$, an initial state $x_0$, and a policy $\pi$, the state evolution $S = \{x_0,x_1,\dots,x_T\}$ is random due to probabilistic state transitions.  Then provided a similar sequence of risk measures $\hat\rho_T = \{\rho_i\}_{i=0}^{T-1}$, we can define a cost function $J$ corresponding to sequential risk evaluations of each cost random variable over the state horizon, with perhaps some discount factor $\gamma \in (0,1]$:
\begin{equation}
\begin{aligned}
    \label{eq:risk-aware-bhp-cost}
    J(x_0,\pi,\gamma) = &\rho_0\bigg(C(x_0,\pi(x_0)) + \gamma \rho_1\big(C(x_1,\pi(x_1)) +  \\
    & \gamma^2 \rho_2(C(x_2,\pi(x_2)) +\dots  \bigg).
\end{aligned}
\end{equation}
Here, we note that while we have defined the interior states $x_1,x_2,\dots,x_{T-1}$, they are random based on the probabilistic transitions from the previous state and the provided action.  As such, this randomness must be accounted for during risk-measure evaluation, resulting in the nested risk measures in the cost above.  Similarly then, the optimal policy minimizes this cost:
\begin{equation} 
\label{eq:MDPproblem}
\pi^* = \arg\min_{\pi}  J(x_0,\pi,\gamma).
\end{equation}
Finally, for $\gamma < 1$ this is called the \textit{discounted MDP problem}, for $\gamma = 1$ this is called the \textit{un-discounted MDP problem}, and for either case where $T \to \infty$ this is called the \textit{infinite-horizon MDP problem}, all of which have been applied to risk-aware behavior planning as will be discussed in the sections to follow.

The last extension on the MDP framework concerns the case where the state is not exactly known, \textit{i.e.} the system is partially observed at every time step, and as such, we do not know the exact state the system occupies.  Rather, we have a belief, a probability distribution, over the states which encodes our understanding of where the system is at any given time.  We'll refrain from rigorously defining this Partially Observable Markov Decision Problem (POMDP) as, at a high level, the policy selection pursuit is similar in spirit to what was shown in the prior two optimization problems (for more information on POMDPs though, please reference~\cite{krishnamurthy2016partially,ahmadi2020control}).  The primary change arises in the consideration of the distribution over states when accounting for risk, whereas the state was known exactly in the prior, MDP case.
% }

\subsubsection{Examples}
Perhaps the most prevalent use of (PO)MDPs in planning arises in navigation problems wherein the finite (PO)MDP states correspond to a discretization of the robot's state space $\mathcal{X}$ into a finite set, \textit{e.g.} discrete cells on a 2-d grid for planar navigation, combined with the enumeration of a finite list of environmental phenomena, \textit{e.g.} the occupancy values of those same grid cells~\cite{ono2015chance}.  Actions then correspond to a minor navigation task, \textit{e.g.} moving from one discretized grid cell in the state space to another discretized grid cell.  Finally, stochasticity between these transitions arises as either the environmental phenomena is partially observed and might frustrate the immediate application of the minor navigation action, faulty sensors causing some drift in the robot's ability to directly achieve the minor navigation task, \textit{etc}.  Then, the goal is to identify an optimal policy against a risk measure, typically VaR or CVaR.  Indeed, this paradigm underlies many works in this vein, a sampling of which is provided here and in the citations within~\cite{nardi2019uncertainty,wohlke2021hierarchies,ono2015chance,ahmadi2021risk,ahmadi2021constrained,wei2022cvar,stegmaier2022cooperative}.  These works are by no means exhaustive, however, and due to the prevalence of MDPs for behavior planning, both control theorists and roboticists alike have also progressed fundamental research in (PO)MDPs in pursuit of advancements in robotic behavioral path planning.  A discussion on those advancements will follow.
}

\newidea{Prior Work on Discounted MDPs} In an early work in this vein, the authors of~\cite{ruszczynski2010risk} presented techniques for incorporating this measure into dynamic programming. This work resulted in a wave of new work evaluating risk measures in dynamic programming problems~\cite{shapiro2021lectures,chow2017risk,majumdar2020should,prashanth2022risk,bauerle2014more}.  For example, in ~\cite{prashanth2014policy,chow2014algorithms} the authors identified locally optimal solutions via gradient descent, to MDP problems with CVaR constraints and total expected costs.  Notably, \cite{prashanth2014policy} provides a convergence guarantee whereas~\cite{chow2014algorithms} does not. 
The authors of~\cite{tamar2016sequential,tamar2015policy} extended these prior notions by developing sample-based saddle point algorithms to identify policies for MDPs whose cost is a coherent risk measure, though not specifically CVaR.  Other relevant works include~\cite{haskell2015convex,feyzabadi2014risk,pereira2013risk,lam2023risk}.

One question posed by the authors in~\cite{hau2023dynamic} has caused a renewal of work in this vein.  Specifically, the authors show that most risk-level dynamic programs cannot guarantee the recovery of a globally optimal value function despite discretized state space.  To partially address that concern, in ~\cite{ahmadi2021constrained,ahmadi2023risk} the authors generate optimal risk-aware policies for MDPs with dynamic coherent risk objectives and constraints.  By phrasing policy generation as a difference convex program, solutions can also be rapidly identified.  Despite these advances, the field of risk-aware discounted MDPs still holds numerous avenues for future exploration. New algorithms and techniques that can handle an expansive range of coherent risk measures and can effectively manage constraints in MDPs are needed. 

\addendum{
\newidea{Prior Work on POMDPs} While POMDPs can be difficult to design and solve, significant strides have been made in addressing coherent risk measure objectives. For instance,~\cite{fan2018risk} explored POMDPs with coherent risk measure objectives. However, their noteworthy theoretical contributions fell short of providing a computational method for designing policies applicable to general coherent risk measures.  Ahmadi et al.~\cite{ahmadi2020risk} aimed to address this gap by proposing a method for finding finite-state controllers for POMDPs with objectives defined in terms of coherent risk measures. Their novel approach took advantage of convex optimization techniques, showcasing the potential of mathematical optimization in policy design. Nevertheless, their method has its limitations: it can only be applied when the risk transition mapping is affine in the policy. 

Recognizing this limitation, Ahmadi et al.\cite{ahmadi2023risk}  extended their prior work~\cite{ahmadi2020risk} to incorporate a broader set of coherent risk measures. They proposed an innovative approach bounded policy iteration method that identifies finite-state risk-averse policies. This methodology breaks the problem down into manageable pieces, tackling convex optimization problems at each policy iteration step. This approach substantially ameliorates the computational tractability of synthesizing risk-averse policies for POMDPs. By iteratively solving these convex optimization problems, the policy synthesis process becomes markedly more feasible. 

However, the methodology outlined in~\cite{ahmadi2023risk} has its limitations. One notable constraint is that the technique can currently only be applied to problems involving hundreds of states due to the computational limitations inherent to convex optimization.
Despite existing limitations, the exploration of POMDPs in the context of coherent risk measures presents a promising field of study. As our understanding deepens and computational methods evolve, we can anticipate the development of more pragmatic solutions for planning under uncertainty under a broader range of coherent risk measures.
}

%% --------------------------------------- %%
\begin{figure}[t] % ’ht’ tells LaTeX to place the figure ’here’ or at the top of the page
\vskip -0.1 true in
\centering
%\resizebox{4cm}{4cm}{
\begin{tikzpicture}
\node[state, initial] (1) {$x^i$};
\node[state, below left of=1] (2) {$x^g$};
\node[state, right of=2] (3) {$x^j$};
\draw (1) edge[above] node{} (2)
(1) edge[below, bend right, left=0.3] node{} (3)
(1) edge[loop above] node{} (1)
(2) edge[loop above] node{$T(\addendum{x^g\mid x^g,u})=1 \quad$} (2)
(3) edge[loop right] node{} (3)
(3) edge[below] node{} (2)
(3) edge[above, bend right, right=0.3] node{} (1);
\end{tikzpicture}
%}
\caption{The transition graph of a transient MDP. The goal state \addendum{$x=x^g$} is cost-free and absorbing. }
\vskip -0.1 true in
\label{fig:fig1}
\end{figure}
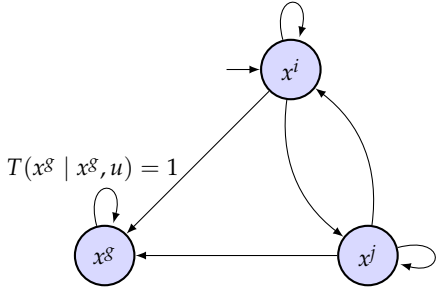

\subsection{Risk-Aware Motion Planning and Control}\label{subsection:motion_planning}
\addendum{The behavior planning layer as described in the previous section is only the top layer of most controllers.  Below this layer, controllers typically comprise of a motion planning layer translating the broken-down behavior into a lower-level sequence of commands which are then tracked by the lower-level controller.  This section will detail risk-aware work done on the remaining two layers - motion planning and control.}
Optimization-based motion planning methods are increasingly popular because they provide optimized system behaviors that respect the system's dynamics, while readily incorporating state and control constraints.
Critically, one must account for sudden system changes or disturbances to ensure system safety.  
Risk-awareness methods can be integrated into optimal planning control problems.

Our risk-aware motion planning review considers discrete-time controlled stochastic systems, with the form:
\begin{equation}\label{eq:sys}
    \boldsymbol{x}(t+1) = f(\boldsymbol{x}(t), \boldsymbol{u}(t), \boldsymbol{d}(t)).
\end{equation}
Here, $\boldsymbol{x}(t) \in \mathbb{R}^{n_x}$ and $\boldsymbol{u}(t) \in \mathbb{R}^{n_u}$ are the system state and controls at time $t$, respectively. The system is affected by a stochastic process noise $\boldsymbol{{d}}(t) \in \mathbb{R}^{n_d}$ and $f:\mathbb{R}^{n_x}\times\mathbb{R}^{n_u}\times\mathbb{R}^{n_d}\rightarrow\mathbb{R}^{n_x}$.  An optimization-based planner seeks to minimize a system cost $J(\boldsymbol{x}, \boldsymbol{u})\in \mathbb{R}$ for initial condition $\boldsymbol{x}(0) = \boldsymbol{x}_0$ at time $t=0$.   
The optimal controller $U = 
[\boldsymbol{u}(0), \dotsc, \boldsymbol{u}(N-1)]$ is the solution to the following optimization problem:  
\begin{subequations}
\begin{align}
     J^*(\boldsymbol{x}(0)) = \min_{U} \quad&\rho\bigg(\sum_{t=0}^{N-1}J(\boldsymbol{x(t)}, \boldsymbol{u(t)})\bigg), \\
\text{s.t.} \quad & \boldsymbol{x}(t+1) = f(\boldsymbol{x}(t), \boldsymbol{u}(t), \boldsymbol{d}(t)), \\
& \boldsymbol{x}(0) = \boldsymbol{x}_0, \quad \forall t \in \{0, \dotsc N-1\}  & 
\end{align}
\label{eq:MPC}
\end{subequations}

\vskip -0.1 true in
\subsubsection{Risk-Aware MPC} 

Model Predictive Control (MPC) applies the finite-horizon controller (\ref{eq:MPC}) in a receding-horizon fashion. 
Uncertainty can arise for many reasons. Uncertainty in the robot's dynamics model causes the true system motion to differ from the predicted one.  Such effects are typically accounted for via process noise, $\boldsymbol{d}(t)$.  Sensor noise and imprecise robot localization or estimation of environment states are other common sources of uncertainty. There are many ways to account for these uncertainties in an MPC framework.  For example, Robust MPC accounts for worst-case disturbances in a set of bounded uncertainties~\cite{bemporad1999robust}. Robust approaches are often too conservative because they focus on worst-case events. Conversely, stochastic MPC~\cite{mesbah2016SMPC} only accounts for the average realization of the cost while respecting a bound on the probability of violating the state and control constraints, see Sidebar~\ref{sb:mpc_sb}. The resulting policy,  which is often too optimistic, minimizes the MPC objective in expectation instead of usefully accounting for events in the tail of the uncertainty distribution. 

Risk-aware MPC optimizes risk-averse behavioral policies: they are not as conservative as in the robust case. But since they account for ``risky'' outcomes in the tail of the uncertainty distribution, they perform better in practice. In~\cite{singh2018framework}, the authors provide an MPC scheme for a discrete-time dynamical system with process noise whose objective was a Conditional Value-at-Risk (CVaR) measure. They further provided new Lyapunov conditions for risk-sensitive exponential stability. In~\cite{dixit2022tvd}, the authors devised an MPC scheme that expressed a distributionally-robust chance constraint along with a risk-aware cost in terms of a CVaR reformulation. Optimal control using distributionally-robust CVaR constraints with second-order moment ambiguity sets is posed as a semidefinite program in~\cite{parys2016DRCC}. A tree-based approach for MPC that enumerates all possible extreme disturbance signals and searches for feedback policies that account for a tradeoff between robustness and performance through CVaR metrics was proposed in~\cite{chen2022interactive}. In~\cite{Sopasakis2019}, the authors considered multistage risk-averse and risk-constrained optimal control for general coherent risk measures with conic representations. \addendum{A sampling-based approach, using model predictive path integral, to solve nonlinear, optimal control problems using CVaR risk constraints was considered in~\cite{yin2023risk}. Other sampling-based approaches to solving the optimal control problem with state estimation uncertainty involve using a robust version of the RRT$^{*}$ algorithm~\cite{lindemann2021robust}. }

Data-driven MPC that uses samples from the uncertainty distribution is becoming increasingly popular. Risk-aware MPC approaches provide the required robustness to account for the gap between enforcing the sample-based chance constraints for the empirical distribution and the true chance constraint for the actual uncertainty distribution. In~\cite{coulson2019data, coulson2021distributionally} the authors propose a distributionally-robust data-enabled predictive control (DeePC) algorithm, that uses finite samples of an unknown system to make trajectory predictions. Instead of learning the system dynamics model, the authors propose an \textit{equivalent} formulation using these data-driven trajectory predictions that enjoys strong out-of-sample guarantees using Wasserstein distributionally-robust CVaR constraints. Reference \cite{zolanvari2022risklmpc} considers a learning MPC framework whose infinite horizon, CVaR-constrained, optimal control solution is approximated iteratively given a finite number of safe states and uncertainty samples. Through this iterative method, the authors construct a data-driven terminal set for distributionally-robust CVaR-constrained iterative MPC with safety and feasibility guarantees.

MPC is also useful for obstacle avoidance in motion planning tasks. Risk-aware MPC accounts for varied obstacle behaviors and sensor and process noise not limited to Gaussian distributions. The MPC scheme in~\cite{hakobyan2019risk} avoids moving obstacles using a CVaR risk metric. Similar results were obtained in~\cite{dixit2020risksensitive} on the Entropic Value-at-Risk (EVaR) metric for obstacle avoidance with additional guarantees of recursive feasibility and finite-time task completion while following a set of waypoints. In~\cite{dixit2022riskaverse}, these results were extended to general coherent risk measures for systems with process noise to obtain a disturbance feedback policy. The authors propose various constraint-tightening techniques to make the risk-aware obstacle avoidance MPC computationally tractable for motion planning \addendum{while providing probabilistic guarantees on recursive feasibility and task completion in finite time}. A risk-constrained MPC formulation was also studied in~\cite{dixit2023step, fan2021step} wherein the authors computed a risk map for traversing over rough terrain using CVaR. They incorporated this CVaR terrain map into MPC constraints to account for obstacles and terrain hazards. \addendum{All of the above methods, including the formulation presented in Sidebar~\ref{sb:mpc_sb}, consider pointwise-in-time constraints for safety. Instead the authors of~\cite{lew2023risk}  consider a CVaR-based optimal control formulation for obstacle avoidance such that the risk constraints hold for the entire trajectory by accounting for the time-wise supremum safety cost inside the CVaR risk.}

\vskip 0.15 true in

\begin{open_prob}[Model Predictive Control with Uncertainty]\label{sb:mpc_sb}
\noindent Consider a linear, discrete-time system given by 
\begin{equation}\label{eq:sys}
    \boldsymbol{x}(t+1) = A\boldsymbol{x}(t) + B\boldsymbol{u}(t) + D\boldsymbol{d}(t)
\end{equation}
where $\boldsymbol{x}(t) \in \mathbb{R}^{n_x}$ and $\boldsymbol{u}(t) \in \mathbb{R}^{n_u}$ are the system state and controls at time $t$, respectively. The system is affected by a stochastic, additive, process noise $\boldsymbol{{d}}(t) \in \mathbb{R}^{n_d}$.

\noindent Consider {$r_x$ state constraints} of the form $$\mathcal{X}\addendum{_S} \coloneqq \{\boldsymbol{x} \in \mathbb{R}^{n_x} | F_x\boldsymbol{x} \leq g_x\}, F_x\in \mathbb{R}^{r_x \times n_x},\, g_x \in \mathbb{R}^{r_x}.$$ 

\noindent We also assume {$r_u$ control constraints} having the form $$\mathcal{U} \coloneqq \{\boldsymbol{u} \in \mathbb{R}^{n_u} | F_u\boldsymbol{u} \leq g_u\}, F_u\in \mathbb{R}^{r_u\times n_x},\, g_u \in \mathbb{R}^{r_u}.$$ 

\noindent The goal is to steer the system to a set
\begin{align*}
    \mathcal{X}_F \coloneqq \{\boldsymbol{x} \in \mathbb{R}^{n_x} | F_f\boldsymbol{x} \leq g_f\}, F_f\in \mathbb{R}^{r_f \times n_x},\, g_x \in \mathbb{R}^{r_f},
\end{align*}
while minimizing the control effort and deviation from the desired trajectory, i.e., we want to minimize the following cost:
\begin{align*}
    J(\boldsymbol{x}, \boldsymbol{u}) \coloneqq \boldsymbol{x}^TQ\boldsymbol{x} + \boldsymbol{u}^TR\boldsymbol{u},
\end{align*}
where $Q \in \mathbb{R}^{n_x\times n_x}$ and $R \in \mathbb{R}^{n_u\times n_u}$ are weights on the state and control costs. 
Model Predictive Control (MPC) provides an optimization-based framework to compute the best $N$-step control input while satisfying the state and control constraints. The MPC optimization is given by,  
\begin{subequations}~\label{eq:SMPC}
\begin{align}
     J^*_{t}(\boldsymbol{x}(t)) = \min_{U_t} \quad&\mathbb{E}\Big[\boldsymbol{x}^T_{t+N|t}P\boldsymbol{x}_{t+N|t} +\\ & \quad \sum_{k=t}^{t+N-1}\big(\boldsymbol{x}^T_{k|t}Q\boldsymbol{x}_{k|t} + \boldsymbol{u}^T_{k|t}R\boldsymbol{u}_{k|t} \big)\Big] \\
\text{s.t.} \quad & \boldsymbol{x}_{k+1|t} = A\boldsymbol{x}_{k|t} + B\boldsymbol{u}_{k|t} + D\boldsymbol{d}_{k|t},  \\
& \text{Prob}(\boldsymbol{x}_{k|t} \not\in \mathcal{X}\addendum{_S}) \leq \beta, \,\, \boldsymbol{u}_{k|t} \in \mathcal{U},\\
& \text{Prob}(\boldsymbol{x}_{t+N|t} \not\in \mathcal{X}_F) \leq \beta \\
& \boldsymbol{x}_{t|t} = \boldsymbol{x}(t) \quad \forall k \in \{t, \dotsc t+N-1\},
\end{align}
\end{subequations}
where, $\boldsymbol{x}_{k|t}$ is the state at time $k$ as predicted at the time $t$ while starting from the current state $\boldsymbol{x}_{t|t}= \boldsymbol{x}(t)$ and $\beta$ is the user chosen risk level.
Uncertainty is propagated through the states as,
\begin{align*}
\boldsymbol{x}_{k+1|t} = A^{k}\boldsymbol{x}_{t|t} + \sum_{i=t}^{k} \big(A^{(k-i)}B\boldsymbol{u}_{i|t} +  A^{(k-i)}D\boldsymbol{d}_{i|t} \big).
\end{align*}
If the uncertainty is i.i.d Gaussian with $\boldsymbol{d}(t) \sim \mathcal{N}(0, \Sigma)$, the states $\boldsymbol{x}(t)$ are also Gaussian $\boldsymbol{x}_{k+1|t} \sim \mathcal{N}\big(\hat{\boldsymbol{x}}_{k|t}, \Sigma_{k|t} \big)$ where, $\hat{x}_{k|t} = A^{k}\boldsymbol{x}_{t|t} + \sum_{i=t}^{k} A^{(k-i)}B\boldsymbol{u}_{i|t}$ and $ \Sigma_{k|t} =\sum_{i=t}^{k} D^T{A^{(k-i)}}^T\Sigma A^{(k-i)}D $ (the family of normal distributions is closed under linear transformations). Hence, we can rewrite the above uncertain MPC optimization as the following deterministic quadratic program,
\begin{align*}
     J^*_{t}(\boldsymbol{x}(t)) = \min_{U_t} \quad&\mathbb{E}\Big[\boldsymbol{x}^T_{t+N|t}P\boldsymbol{x}_{t+N|t} + \\ & \quad \sum_{k=t}^{t+N-1}\big(\boldsymbol{x}^T_{k|t}Q\boldsymbol{x}_{k|t} + \boldsymbol{u}^T_{k|t}R\boldsymbol{u}_{k|t} \big)\Big] \\
\text{s.t.} \quad & \boldsymbol{x}_{k+1|t} = A\boldsymbol{x}_{k|t} + B\boldsymbol{u}_{k|t} + D\boldsymbol{d}_{k|t},  \\
& F_{x}\hat{\boldsymbol{x}}_{k|t} + F_x\Phi^{-1}(1-\beta)\Sigma_{k|t}\leq g_x,\\
& F_u\boldsymbol{u}_{k|t} \leq g_u, \\
& F_{f}\hat{\boldsymbol{x}}_{t+N|t} + F_f\Phi^{-1}(1-\beta)\Sigma_{t+N|t}\leq g_f\\
& \boldsymbol{x}_{t|t} = \boldsymbol{x}(t) \, \forall k \in \{t, \dotsc t+N-1\}.
\end{align*}
However, if the uncertainty distribution is non-Gaussian, the uncertain MPC~\eqref{eq:SMPC} is not easily reformulated into a convex optimization program. In this case, \addendum{one possible solution is to sample }the uncertainty distribution and reformulate the MPC optimization as a much more computationally expensive mixed-integer program~\cite{blackmore2006probabilistic, blackmore2010particle}. Many tail risk measures such as CVaR and EVaR, provide intuitive convex, inner approximations of chance constraints regardless of the uncertainty distribution. Hence, we propose risk-aware MPC formulations not only better account for uncertainty but also provide an efficient convex reformulation without making assumptions about the nature of the uncertainty. The resulting deterministic, risk-aware MPC formulation is given by,
\begin{subequations}
\begin{align*}
     J^*_{t}(\boldsymbol{x}(t)) = \min_{U_t} \quad&\rho\Big[\boldsymbol{x}^T_{t+N|t}P\boldsymbol{x}_{t+N|t} + \\ & \sum_{k=t}^{t+N-1}\big(\boldsymbol{x}^T_{k|t}Q\boldsymbol{x}_{k|t} + \boldsymbol{u}^T_{k|t}R\boldsymbol{u}_{k|t} \big)\Big] \\
\text{s.t.} \quad & \boldsymbol{x}_{k+1|t} = A\boldsymbol{x}_{k|t} + B\boldsymbol{u}_{k|t} + D\boldsymbol{d}_{k|t},  \\
& \rho_{\beta}(F_x\boldsymbol{x}_{k|t} - g_x) \leq 0, \,\, \boldsymbol{u}_{k|t} \in \mathcal{U},\\
& \rho_{\beta}(F_f\boldsymbol{x}_{t+N|t} - g_f) \leq 0 \\
& \boldsymbol{x}_{t|t} = \boldsymbol{x}(t) \quad \forall k \in \{t, \dotsc t+N-1\}.
\end{align*}
% \label{eq:RMPC}
\end{subequations}

\end{open_prob}

\subsubsection{Risk-Aware Safety-Critical Control}

\addendum{In the previous subsections, we considered behavior and motion planning algorithms through the lenses of a risk-aware approach. Risk-aware feedback control methodologies, like the MPC techniques discussed in the previous subsection, can strike a balance between worst-case and nominal operating conditions and thus account for rare events by enhancing robustness while achieving high performance under nominal operation. Additionally, we must also consider risk with a view to safety-critical autonomous systems, such as those found in aerospace and human-robot applications. The specific risks here are often associated with the uncertainty of modeling intricate nonlinear dynamics, e.g. bipedal robots~\cite{reher2020dynamic}, and/or sensing extreme unstructured environments, e.g. subterranean or extraterrestrial exploration~\cite{rouvcek2019darpa}. Safety in the feedback control layer is often formulated in terms of set-theoric properties of dynamical systems~\cite{blanchini2008set}, e.g., reachability and invariance.} Safety verification then involves ensuring that system solutions stay within a predefined safe set or, conversely, steer clear of a predetermined unsafe set. A common technique for this is to calculate the reachable set of a system under disturbances and controls~\cite{abate2008probabilistic,althoff2008reachability,mitchell2005time}. Yet, for intricate, high-dimensional systems, these methods may be impractical or excessively conservative. 

Historically, alternative methods for assessing reachability trace back to Nagumo's seminal research~\cite{nagumo1942lage} on the set invariance of ordinary differential equations (ODEs). This work was later expanded to include ODEs with inputs by Aubin and others, under the framework of viability theory~\cite{aubin2011viability}. The rise of interest in hybrid systems in the 2000s led to the development of barrier certificates for safety verification~\cite{prajna2007framework}. The creation of these certificates, however, involves solving complex polynomial optimization problems that are challenging for high-dimensional systems, despite some progress made in the last decade~\cite{ahmadi2014dsos}. The newer concept of barrier functions~\cite{ames2016control} offers a solution to the computational difficulties faced by barrier certificates. These functions can be formulated directly from the safe set's definition, simplifying the process. Utilizing this attribute, barrier functions have been effectively applied to design safe controllers (without an existing controller) and safety filters (with an existing controller) for dynamic systems like biped robots~\cite{nguyen20163d} and trucks~\cite{chen2019enhancing}. These applications have demonstrated assured performance and robustness~\cite{xu2015robustness}.

Conditional Value-at-Risk is a useful measure for assessing how far a realized trajectory may deviate from a safe region of operation~\cite{parys2016DRCC,SamuelsonInsoon2018, chapman2019risk, chapman2021risk, chapman2022optimizing, weifausschapmanconf2022}. 
Defining safety in terms of CVaR is well-motivated when constraint violations may be unavoidable: the magnitude of the risk should be minimized during the undesired excursion  %minimized or at least penalized
\cite{parys2016DRCC, chapman2022optimizing}.  Sets of initial conditions whose safety is characterized by motions of CVaR can be estimated using dynamic programming~\cite{SamuelsonInsoon2018, chapman2019risk, chapman2021risk, chapman2022optimizing, weifausschapmanconf2022}. Pointwise CVaR constraints have also been studied in~\cite{SamuelsonInsoon2018}. The problem of optimizing the CVaR of a maximum random cost has been studied in different settings, such as deriving an upper bound approximation~\cite{chapman2021risk}, a finite-horizon solution~\cite{chapman2022optimizing}, and an infinite-horizon solution~\cite{weifausschapmanconf2022}.

Safety requirements can also be encoded and enforced via Control Barrier Functions (CBFs), which were proposed in~\cite{ames2016control}. CBFs have been used to design safe controllers for continuous-time dynamical systems, such as bipedal robots~\cite{nguyen20163d}
and trucks~\cite{chen2019enhancing}, with guaranteed robustness~\cite{xu2015robustness,kolathaya2018input} (see the survey~\cite{ames2019control} and references therein).  For discrete-time systems, discrete-time barrier functions were formulated in~\cite{ahmadi2019safe,agrawal2017discrete} and applied to multi-robot coordination~\cite{ahmadi2020barrier}. \addendum{For a class of stochastic (Ito) differential equations, safety in probability and statistical mean was studied in ~\cite{so2023almost,santoyo2019barrier,yaghoubi2020risk,yaghoubi2021risk,vahs2023belief} via stochastic barrier functions.}

The first attempt to formulate risk-aware control barrier functions was carried out in~\cite{ahmadi2021riskbipedal}, wherein the authors proposed CVaR control barrier functions as a composition of a dynamic CVaR metric with a CBF to study safety, in the CVaR sense, for a discrete-time dynamical system subject to stochastic uncertainty. A computational method based on difference convex programs (DCPs) was also proposed in order to synthesize CVaR-safe controllers. The method was applied to collision avoidance scenarios involving a bipedal robot subject to modeling uncertainty. The CVaR control barrier functions were generalized to risk-aware control barrier functions (RCBFs) with general coherent risk measures in~\cite{singletary2022safe,singletary2023controlling}, where it was shown that the existence of such barrier functions implies invariance
in a coherent risk sense. Furthermore, conditions were proposed based on finite-time RCBFs to guarantee finite-time reachability to a desired set. In recent work \cite{vahs2023risk}, sampling-based under-approximations of the CVaR for belief states were used to define risk CBFs. 

\begin{open_prob}[Risk-Aware Control Barrier Functions]\label{sb:mpc_risk_cbf}
\noindent Consider a discrete-time stochastic system  given by
\begin{equation}\label{eq:dynamics}
    \boldsymbol{x}(t+1) = f\left(\boldsymbol{x}(t),\boldsymbol{u}(t),\boldsymbol{d}(t)\right), \quad x(0)=x_0,
\end{equation}
where at time $t \in \mathbb{N}_{\ge 0}$, $\boldsymbol{x}(t) \in \mathcal{X} \subset \mathbb{R}^n$ is the state, $\boldsymbol{u}(t) \in \mathcal{U} \subset \mathbb{R}^m$ is the control input,  $\boldsymbol{d}(t) \in \mathcal{D}$ is the stochastic uncertainty/disturbance, and $f:  \mathbb{R}^n \times \mathcal{U} \times \mathcal{D} \to \mathbb{R}^n$. We assume that the initial condition $\boldsymbol{x}_0$ is deterministic and that $|\mathcal{D}|$ is finite, \textit{i.e.,} $\mathcal{D} = \{v_1, \ldots, v_{|\mathcal{D}|}\}$. At every time step $t$,
for a state-control pair $(\boldsymbol{x}(t), \boldsymbol{u}(t))$, the process disturbance $\boldsymbol{d}(t)$ is
drawn from set $\mathcal{D}$ according to the probabilities $p = [p_1,\dots,p_{|\mathcal{D}|}]^T$, where $p_i\coloneqq\mathbb{P}(\boldsymbol{d}(t)=v_i)$, $i=1,2,\ldots,|\mathcal{D}|$. Note that the
probability mass function for the process disturbance is time-invariant, and that the process disturbance is independent of
the process history and of the state-control pair $(\boldsymbol{x}(t), \boldsymbol{u}(t))$. Note that, in particular, system~\eqref{eq:dynamics} can capture stochastic hybrid systems, such as Markovian Jump Systems. 

In risk-aware safety analysis, we are interested in studying the properties of the solutions to~\eqref{eq:dynamics} with respect to the compact set $\mathcal{S}$ described by: 
\begin{align}
\label{eq:safeset}
\mathcal{S} \coloneqq\{ \boldsymbol{x} \in \mathcal{X} \mid h(\boldsymbol{x}) \ge 0 \}, \nonumber\\
\mathrm{Int}(\mathcal{S}) \coloneqq\{ \boldsymbol{x} \in \mathcal{X} \mid h(\boldsymbol{x}) > 0 \}, \\
\partial \mathcal{S} \coloneqq\{ \boldsymbol{x} \in \mathcal{X} \mid h(\boldsymbol{x}) = 0 \}, \nonumber
\end{align}
where $h:\mathcal{X} \to \mathbb{R}$ is a continuous function.

In the presence of stochastic uncertainties $\boldsymbol{d}$, assuring almost sure (with probability one) invariance or safety may not be feasible. Moreover, enforcing safety in expectation is only meaningful if the law of large numbers can be invoked and we are interested in the long-term performance, independent of the actual fluctuations. RCBFs focus on safety in the dynamic coherent risk measure sense with conditional expectation as a special case, allowing for more robust measures of safety.

%\vspace{0.1cm}
\begin{defin} [$\rho$-Safety]
Given a "safe set" $\mathcal{S}$ in~\eqref{eq:safeset} and a time-consistent, dynamic coherent risk measure $\rho_{0:t}$, the solutions to~\eqref{eq:dynamics}, starting at $\boldsymbol{x}_0 \in \mathcal{S}$, are \emph{$\rho$-safe} if and only if 
\begin{align}\label{eq:risksafety}
    \rho_{0,t}\left(0,0,\ldots, h(\boldsymbol{x}(t)) \right) \ge 0, \quad \forall t \in \mathbb{N}_{\ge 0}.
\end{align}
\end{defin}

When $\boldsymbol{x}_0 \in \mathcal{X}\setminus \mathcal{S}$, we often want to know if $\mathcal{S}$ can be reached in finite time.
%\vspace{0.1cm}

\begin{defin}[$\rho$-Reachability]
Consider system~\eqref{eq:dynamics} with initial condition $\boldsymbol{x}_0 \in \mathcal{X} \setminus \mathcal{S}$. Given a set $\mathcal{S}$ as in~\eqref{eq:safeset} and a time-consistent, dynamic coherent risk measure $\rho_{0:t}$, $\mathcal{S}$ is \emph{$\rho$-reachable} if and only if there exists a constant $t^*$ such that
\begin{equation}\label{eq:ftrisksafety}
    \rho_{0,t^*}\left(0,0,\ldots, h(\boldsymbol{x}(t^*)) \right) \ge 0.
\end{equation}
\end{defin}
\end{open_prob}

% \subsection{Risk-Aware Safety with RCBFs}
%\vspace{0.1cm}
\begin{defin}[Risk-Aware Control Barrier Function]\label{def:riskbf}
For~the discrete-time system~\eqref{eq:dynamics} and a dynamic coherent risk measure $\rho$, the continuous function $h : \mathbb{R}^n \to \mathbb{R}$ is a \emph{risk-aware control barrier
function} (RCBF) for the set $\mathcal{S}$ as defined in~\eqref{eq:safeset}, if there exists a convex class-$\mathcal{K}$ function $\alpha$\footnote{A class-$\mathcal{K}$ function is a continuous, scalar function $\alpha(r)$ defined for $r\in [0, a)$ that is strictly increasing  and satisfies $\alpha(0) = 0$.} satisfying $\alpha(r) < r$ for all $r>0$  such that
\begin{equation}\label{eq:BFinequality}
   \rho\left( h(\boldsymbol{x}({t+1)})\right) \ge  \alpha( h\left(\boldsymbol{x}(t))\right),\quad \forall \boldsymbol{x}(t) \in \mathcal{X}.
    \end{equation}
\end{defin}
%\vspace{0.1cm}

% \newsec{Risk Sensitive Safety with RCBFs:}  
In~\cite{singletary2022safe}, the authors demonstrated that the existence of an RCBF implies invariance/safety in the coherent risk measure.

%\vspace{0.1cm}
\begin{theorem}\label{thm:riskbf}
For discrete-time system~\eqref{eq:dynamics} and the set $\mathcal{S}$ as described in~\eqref{eq:safeset}, let $\rho$ be a coherent risk measure. Then, $\mathcal{S}$ is $\rho$-safe if there exists an RCBF as defined in Definition~\ref{def:riskbf}.
\end{theorem}

Note that the most common choice for function $\alpha$ is a constant $\alpha=\alpha_0$, where $\alpha_0 \in (0,1)$, as $\alpha_0 r < r$, $\forall r>0$. To study risk-aware reachability, we require the following.

%\vspace{0.1cm}
\begin{defin}[Finite-Time RCBF] \label{def:ftdtbf}
For  discrete-time system~\eqref{eq:dynamics} and dynamic coherent risk measure $\rho$, the  continuous function ${h}:\mathcal{X} \to \mathbb{R}$ is a \emph{finite-time RCBF} for set $\mathcal{S}$, as defined in~\eqref{eq:safeset}, if there exist constants $0<\gamma<1$ and $\varepsilon > 0$ such that
\begin{equation}\label{eq:BFft}
    \rho\left({h}(\boldsymbol{x}(t+1))\right)-\gamma {h}(\boldsymbol{x}(t)) \ge \varepsilon (1-\gamma),\quad \forall {  \boldsymbol{x}(t) \in \mathcal{X}}.
\end{equation}
\end{defin}
It was also shown in~\cite{singletary2022safe} that the existence of a finite-time RCBF implies $\rho$-reachability.

\begin{theorem}~\label{thm:FTDTBF}
Consider the discrete-time system~\eqref{eq:dynamics} and a dynamic coherent risk measure $\rho$. Let $\mathcal{S} \subset \mathcal{X}$ be as described in~\eqref{eq:safeset}. If there exists a finite-time RCBF ${h}:\mathcal{X}\to \mathbb{R}$ as in Definition~\ref{def:ftdtbf}, then for all $\boldsymbol{x}(0) \in \mathcal{X}\setminus \mathcal{S}$, there exists a $t^* \in \mathbb{N}_{\ge 0}$ such that $\mathcal{S}$ is $\rho$-reachable, \textit{i.e.,} inequality~\eqref{eq:ftrisksafety} holds. Furthermore, 
\begin{equation} \label{eq:upperboundtheorem2}
    t^* \le {\log\left(\frac{\varepsilon - {h}\left(\boldsymbol{x}(0)\right)}{\varepsilon}\right)}/{\log\left(\frac{1}{\gamma}\right)},
    \end{equation}
    where the constants $\gamma$ and $\varepsilon$ are as defined in Definition~\ref{def:ftdtbf}.
\end{theorem}

\subsection{Case Study: Risk-Aware Robotic Motion Planning in Subterranean Environments}  This case study looks at a hierarchical risk-aware {\em traversability} and planning methodology that can be used for autonomous robot (legged or wheeled robot) traversal over extreme terrain~\cite{dixit2023step, fan2021step}, as motivated by the DARPA Subterranean challenge.  \addendum{This is the first framework to integrate a CVaR-based risk-aware planning and control pipeline onto a fully autonomous robotic system. We briefly describe the risk-aware traversability and planning pipeline below.}

We first need to interpret which parts of the environment the robot can traverse.  Evaluation of a natural terrain's \textit{traversability} is difficult due to uncertainties arising from sensor noise and robot localization errors. Furthermore, there are multiple sources of terrain hazards such as steep slopes, loose surface material, sudden elevation drops, and physical obstacles. To account for these different sources of uncertainty systematically, we evaluate the conditional value-at-risk of the terrain hazards to obtain a {\em risk map} that can be used in the planning and control pipeline. The traversability estimate is given by the random variable $R$ that \addendum{is constructed jointly from the grid map of the terrain}, the robot state, and the applied control, \addendum{see~\cite{fan2021step, dixit2023step} for details on how to construct the map and associated traversability estimate. $R$ provides a cost of traversing over each grid point on the terrain that we use to assess the CVaR value.}
This CVaR risk evaluation,  $\mathrm{CVaR}_{\beta}(R)$, enables a robot engineer to define the allowable traversability risk level based on the mission criteria. Furthermore, one can dynamically adjust the risk level, $\beta$, online based on 1) the mission-level states, i.e., based on the robot's capabilities and the environment, and 2) whether the robot is stuck in a situation wherein there is no feasible path and decreasing the risk-level (and consequently being less conservative) might allow the robot to find a feasible, but possibly riskier path. The geometric planner and the kinodynamic MPC controller then utilize the risk evaluation, $\mathrm{CVaR}_{\beta}(R)$, to obtain a risk-aware control policy. 

\begin{figure}
    \centering
    \includegraphics[width = 0.98\columnwidth]{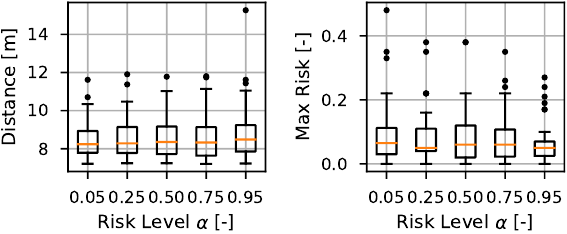}
    \caption{\addendum{\textbf{Left:} Trade-off between the distance traversed by the robot and different risk levels.  \textbf{Right:} the trade-off between maximum risk taken along the traversed path and different risk levels. Note that $\beta = 1 - \alpha$. Figure taken from~\cite{dixit2023step}. }}
    \label{fig:subt_sim_tradeoff}
\end{figure}

\begin{figure}
    \centering
    \includegraphics[width = 0.98\columnwidth]{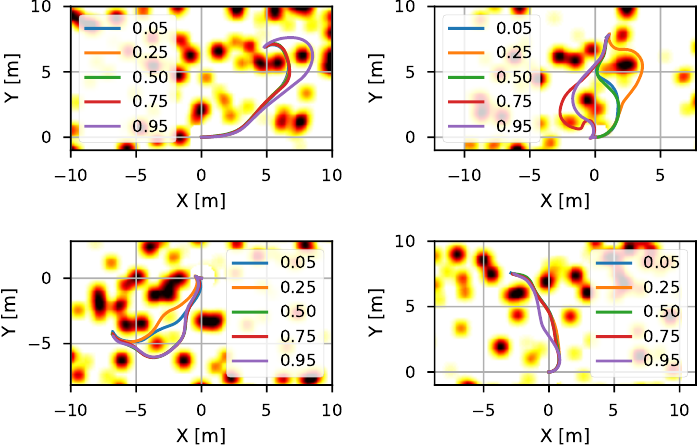}
    \caption{\addendum{Four instances from a  Monte-Carlo simulation illustrate how different choices of risk levels, $1-\beta$, affect the paths taken by the robot. Figure taken from~\cite{dixit2023step}.}}
    \label{fig:subt_sim_path}
\end{figure}

\addendum{After computing the risk-aware traversability cost, the authors utilize this cost for high-level geometric path planning using an A* algorithm to obtain waypoints for navigation. This risk-aware geometric planner uses the dynamic risk measure in the cost of the geometric planner optimization. The waypoints are then passed into an MPC optimization with CVaR constraints for safety for generating obstacle-free trajectories. }The statistical performance of the aforementioned risk-aware controllers is evaluated in simulation with randomly generated environment maps and goals. This study illustrates the trade-off between the risk taken by the robot to reach the goal versus the total distance traversed by the robot for different allowable risk levels, $\beta$ (see Figure~\ref{fig:subt_sim_tradeoff}). \addendum{When $\beta \rightarrow 1$ (or $\alpha \rightarrow 0$), the framework is risk-neutral and mimics the deterministic setting, where we only use the mean value of the traversability estimate. We can see from Figure~\ref{fig:subt_sim_tradeoff}) that the robot takes paths with moderate to high risk more often (i.e. paths with maximum risk  $> \sim0.3$). Ultimately, the risk level desired is left up to the user, but the main takeaway is that the user can change the risk level to mimic deterministic or robust baselines as desired. }The robot uses longer, low-risk paths when the robot is risk-averse (low $\beta$) and shorter, higher-risk paths when the robot is risk-neutral (high $\beta$), see Figure~\ref{fig:subt_sim_path}.

\begin{figure*}
    \centering
\begin{subfigure}[b]{0.99\textwidth}
    \centering
    \includegraphics[width=0.99\textwidth]{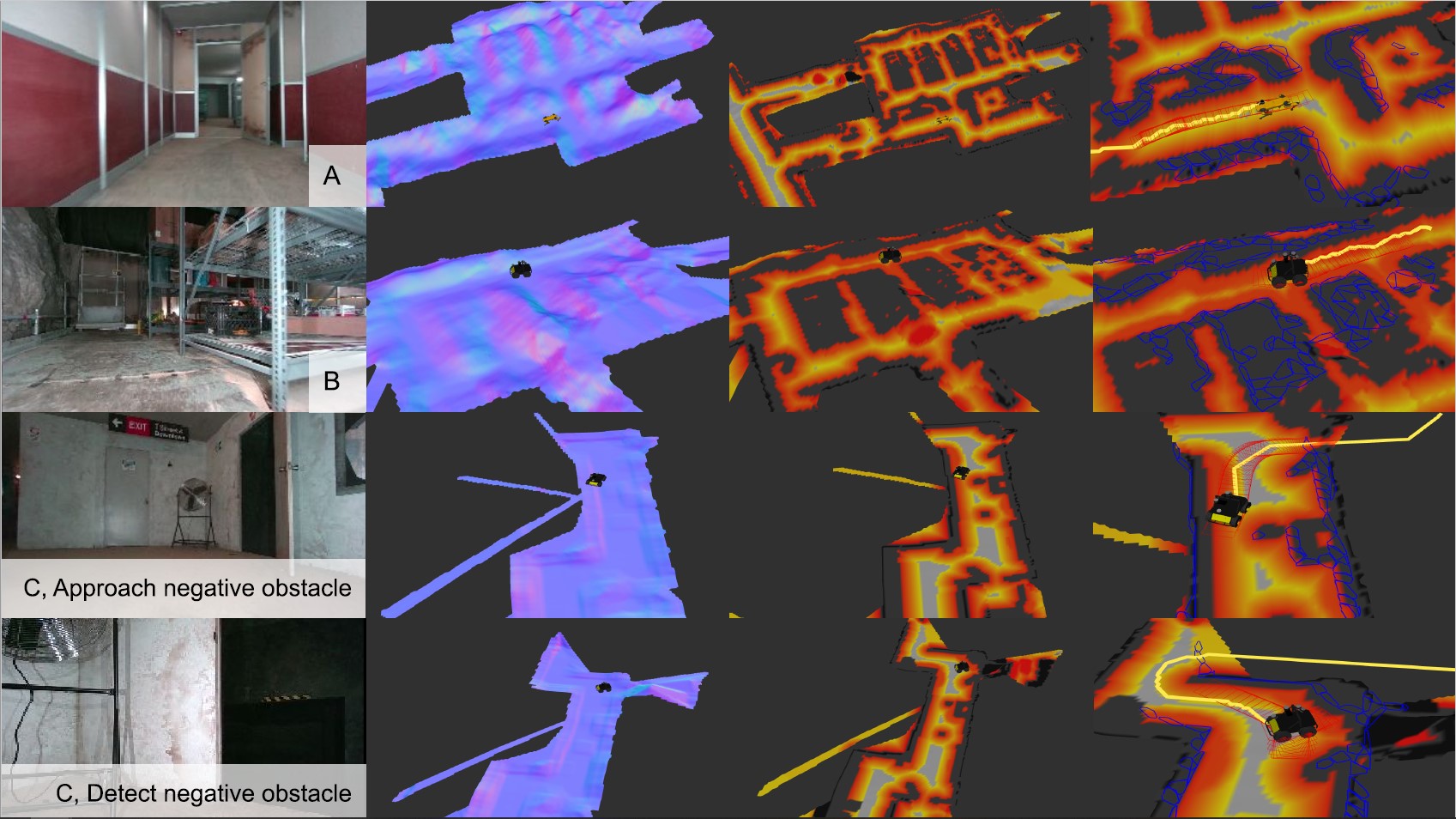}
    % \label{fig:finals_results2}
\end{subfigure}\\
\begin{subfigure}[b]{0.99\textwidth}
    \centering
    \vspace{0mm}
    \includegraphics[width=0.99\textwidth]{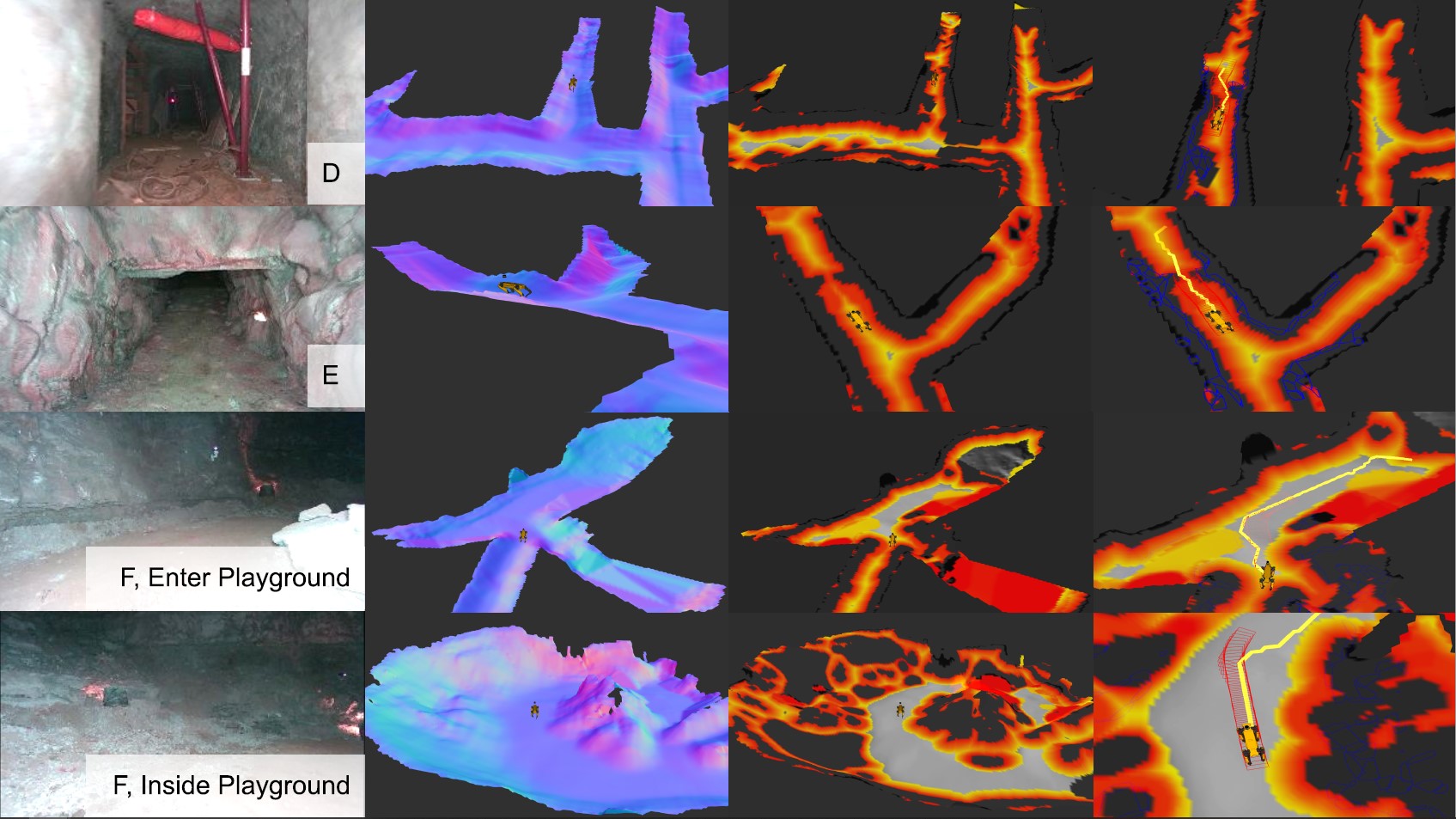}
    % \label{fig:finals_results2}
\end{subfigure}
 \caption{Risk-aware traversability analysis of the DARPA Subterranean Challenge final course. Columns, from left to right: robot front camera view, elevation map (colored by normal direction), risk map (colored by risk level  - white: safe, yellow to red: moderate, black: risky), and planned geometric/kinodynamic paths (yellow lines/red boxes). Figure taken from~\cite{dixit2023step}.}
    \label{fig:finals_results}
\end{figure*}

This risk-aware traversability evaluation and planning framework was experimentally tested during the DARPA Subterranean Challenge and in other real-world subterranean environments.  The final competition course of the DARPA Subterranean Challenge was comprised of tunnel, urban, and cave environments for which the traversability evaluation and navigation results are provided in Figure ~\ref{fig:finals_results}.  The following list describes the difficult terrain hazards found within that environment:
\begin{itemize}
\setlength{\itemindent}{0.5in}
    \item[\textbf{Region A}] An office-like area with narrow corridors and small rooms where it is tough to find a feasible path if the maps are overinflated to avoid obstacles. 
    \item[\textbf{Region B}] A mock post-earthquake warehouse whose shelving and clutter are difficult to navigate around. 
    \item[\textbf{Region C}] A door connecting the urban and tunnel part of the course via stairs. The stairs act as a potential sudden drop-off (\textit{i.e.}, a negative obstacle) for wheeled robots. The drop is hard to detect because of the narrow doorway.
    \item[\textbf{Region D}] A narrow passage littered with debris, vertical pipes along the walls, and ceiling obstacles. The robot must correctly identify the pipes as obstacles.
    \item[\textbf{Region E}] A small cave opening that mimics real caves, wherein humans must crawl through the small openings to reach another cave chamber. The upward-sloping cave floor and downward-sloping ceiling make it difficult to differentiate between the ceiling and the ground. The ceiling height at the opening is very close to the ground height at the end of the opening. 
    \item[\textbf{Region F}] A small limestone cave with rubble and loose rock piles. The robot must distinguish between traversable and non-traversable rubble.
\end{itemize}

A statistical analysis of the simulations and the \addendum{qualitative} experimental results from the field show that a risk-aware traversability and planning pipeline provides a framework where the risk of the \textit{entire} system can be adjusted by changing the risk-level $\beta$ despite there being multiple risk sources, such as slopes, obstacles, low-ceilings, and mud. This framework is agnostic to the kind of ground robot utilized: it has been tested on wheeled robots (Clearpath Husky) and legged robots (Boston Dynamics Spot quadruped).
\begin{figure}[t]
      \centering
      \includegraphics[width = \columnwidth]{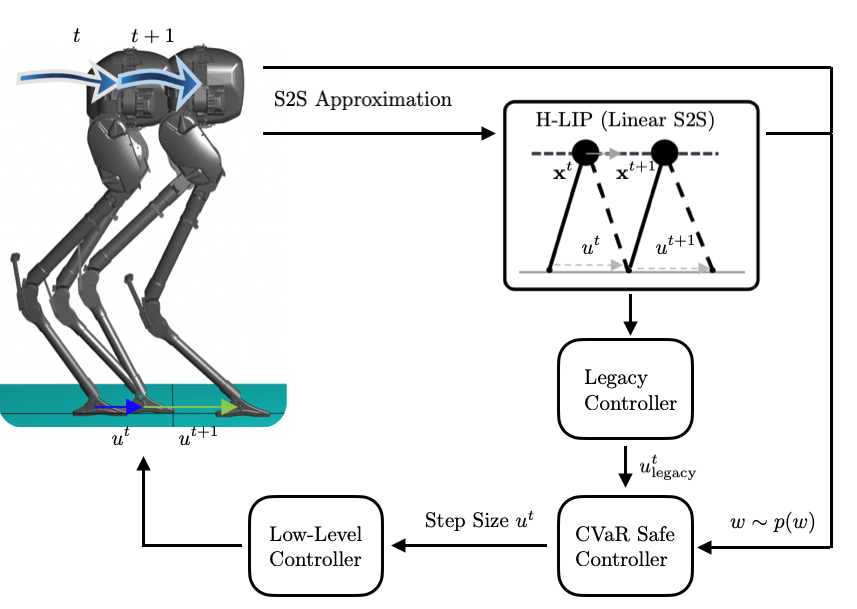}%  
      \caption{{ Schematic diagram of a risk-aware bipedal robot path planning method based on RCBFs with CVaR risk measure~\cite{ahmadi2021riskbipedal}. The diagram features the sequential flow of processes starting with the S2S Approximation of robot dynamics, which leverages the dynamics of the walking robot, modeled after the Hybrid-Linear Inverted Pendulum (H-LIP) system, where $\boldsymbol{x}^t \equiv \boldsymbol{x}(t)$ denotes the horizontal position of the center of mass (COM) of the robot relative to the inertia frame. The H-LIP approximation is controlled via a legacy controller, such as a model predictive control, which outputs step size commands $u_{legacy}$.   The difference in terms of the horizontal position of the COM  between the H-LIP model and simulation/experimental is measured offline and used to construct the discrete distribution over the uncertainty $p(w)$.
      The distribution over uncertainty is used to tune a CVaR-safe controller based on RCBFs. The outputs of the legacy controller are then amended online using the CVaR safe controller to ensure risk-aware safety in the presence of uncertainty $p(w)$, which adjusts the robot's locomotion parameters (in particular, step size $u$) in real-time. }}
      \label{fig:diagram}
\end{figure}
\subsection{Case Study: Risk-Aware Bipedal Walking} Control of bipedal walking presents significant challenges, as evidenced by the variety of approaches taken in the literature to handle the nonlinearity and complexity of bipedal robot dynamics ~\cite{grizzle2014models}. In practice, bipedal walking dynamics are often simplified by approximate models subject to stochastic uncertainty~\cite{xiong2020ral}.  The horizontal robot state at time $t$ is denoted by $\boldsymbol{x}_h(t) = [c, p, v]^T$, where $c$ represents the horizontal position of the robot's center of mass (COM) relative to an inertial frame, $p$ denotes the horizontal COM position with respect to the stance foot, and $v$ denotes the horizontal COM velocity.
\addendum{The step-to-step (S2S)  dynamics of the horizontal COM state is expressed as $\boldsymbol{x}_h(t+1)= \mathcal{P}^h(\boldsymbol{x}(t), \tau(t))$, where $\tau$ represents the joints' input torques.} In practice, deriving the S2S dynamics analytically is challenging due to the robot's nonlinear and hybrid dynamics.

The authors in~\cite{xiong2020ral,xiong2021inReview} suggested that a Hybrid-Linear Inverted Pendulum (H-LIP) walking model ~\cite{xiong2021inReview} provides an apt approximation for the actual horizontal S2S dynamics of robot walking. The H-LIP dynamics are represented as 
\begin{equation}
    \boldsymbol{x}_{\text{H-LIP}}(t+1)= A \boldsymbol{x}_{\text{H-LIP}}(t)+ B \boldsymbol{u}_{\text{H-LIP}}(t),
\end{equation} 
where $\boldsymbol{x}_{\text{H-LIP}}(t+1)= [c_{\text{H-LIP}}, p_{\text{H-LIP}}, v_{\text{H-LIP}}]^T$ is the H-LIP's discrete pre-impact state, and $\boldsymbol{u}_{\text{H-LIP}}(t)$ denotes the step size. The specific expressions for $A$ and $B$ can be found in~\cite{xiong2020ral}. With this approximation, the S2S dynamics can be rewritten as
\begin{equation}
    \boldsymbol{x}_h(t+1) = A \boldsymbol{x}_h(t) + B \boldsymbol{u}(t) + \boldsymbol{d}(t),
\end{equation}
where $\boldsymbol{d}(t):= \mathcal{P}^h(x(t), \tau(t)) - A \boldsymbol{x}_h(t) -B \boldsymbol{u}(t) \in \mathcal{D}$ can be viewed as a stochastic disturbance to the linear system. 

As seen in Figure~\ref{fig:diagram}, a CVaR RCBF-based controller can ensure safe, risk-aware 3D bipedal walking.  The model discrepancy $w$ is treated as a stochastic uncertainty and risk factor that could elicit undesirable walking behavior. To mitigate this risk, CVaR RCBF-based controllers are synthesized and act as safety filters for the H-LIP-based stepping controller.  \addendum{We will omit how these barrier functions were generated for this specific problem to keep the discussion at a high level.  For those more interested in the details, please reference~\cite{ahmadi2021riskbipedal}.  Abstractly, however, these barrier functions delineate safe regions within which the bipedal robot can navigate by constraining the horizontal position of the robot to lie between the obstacles it is navigating between.  To construct this CVaR RCBF, the authors carried out extensive simulations over a wide range of walking behaviors to numerically define an uncertainty set.  Then to calculate the CVaR of the RCBF evaluation at the subsequent time steps - as required in the CVaR RCBF condition expressed in~\cite{ahmadi2021riskbipedal} - the authors approximated the true distribution via a uniform sampling over the previously calculated uncertainty set, and calculated the CVaR against this distribution.  Even still, the corresponding walking experiments still represented safe behavior, further supporting the use of risk-aware control barrier functions for risk-aware safe controller synthesis.}

\begin{figure}[t] \centering{
\includegraphics[scale=.25]{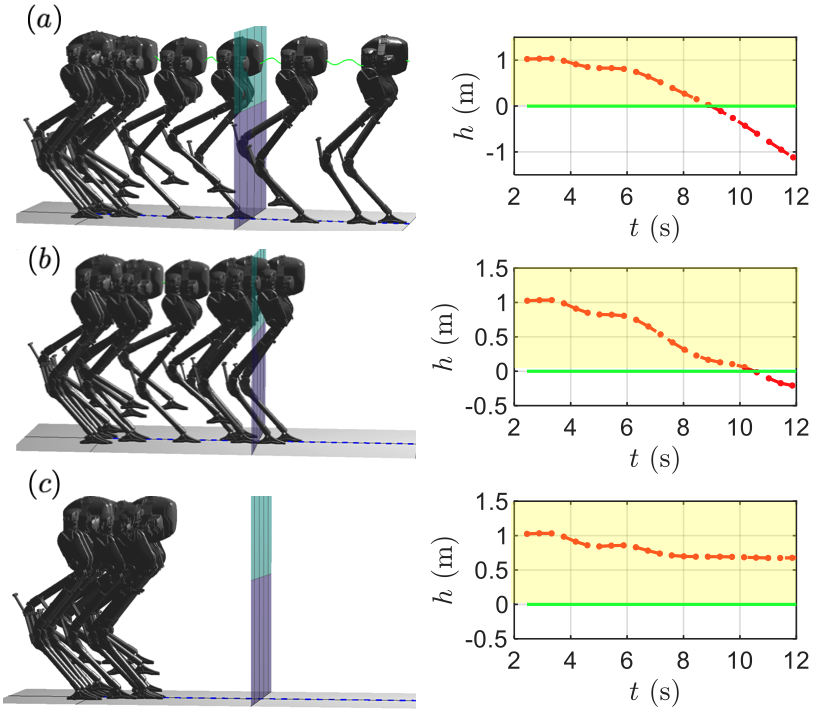}
% \vspace{-.3cm}
\caption{Risk-averse obstacle avoidance using CVaR barrier functions (robot behavior and barrier function evolution). The shaded yellow area denotes safe regions. (a) safety violation with no barrier function; (b) safety violation with risk-neutral barrier function; (c) safe behavior with CVaR barrier function. Plots on the right side show the values of the barrier functions~\cite{ahmadi2021riskbipedal}.}\label{fig:fixedwall}
}
 \end{figure}

\begin{figure*}
    \centering
    \includegraphics[width=\textwidth]{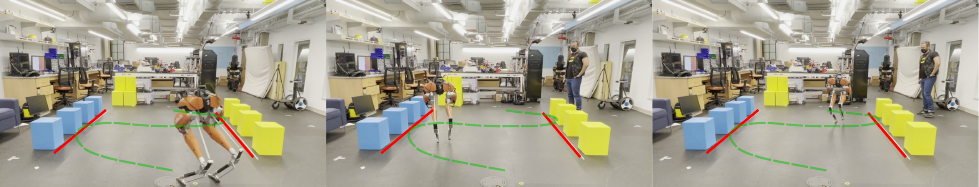}
    \caption{Snapshots from an experiment conducted at Caltech's AMBER Lab, featuring Agility Robotics' Cassie Robot. The OptiTrack system that was used for localization tracks reflective markers on the robot to determine its position and orientation with high precision. The data from OptiTrack was fed into the robot's control system in real-time, allowing it to make immediate adjustments to its path. The code that enables the robot to perform motion planning is executed on a computer embedded within the robot. (left) The initial setup, where the robot's trajectory is aligned with a sinusoidal path, represented by a dashed green line. (middle) Mid-course navigation highlighting the effectiveness of the risk control barrier function-based safety filter. This filter is designed to allow the robot to dynamically avoid obstacles and unsafe regions, which are indicated by the red solid lines, representing the boundaries of areas the robot should not enter. (right) successfully following a trajectory that has been adjusted by the risk-aware safety filter. This demonstrates the practical application of the risk-aware control method outlined in Figure 7, where a robot not only plans its path in consideration of potential risks but also dynamically adjusts its course in real time to maintain safe navigation.}
    \label{fig:cassie-experiment}
    \vskip -0.1 true in
\end{figure*}

The simulation results seen in Figure~\ref{fig:fixedwall} showcase the effectiveness of the CVaR-based CBF, especially when taking into account the inherent uncertainties in 3D bipedal dynamics. Figure ~\ref{fig:cassie-experiment} presents snapshots from an experiment conducted at the Caltech AMBER Lab using the Agility Robotics' Cassie bipedal Robot.

\subsection{Open Questions and Future Directions}
This section outlined recent advances in risk-aware planning and control. We applied these ideas to bipedal walking and terrain traversability analysis for wheeled and legged robots. Many future research directions are suggested by current work in risk-aware planning and control.

%\subsubsection{Computation} 
\textbf{Computation.} The best choice of a risk measure for a specific problem remains an open question. The popular CVaR risk measure is computationally attractive as it can be formed into linear programs \addendum{when the original stochastic optimal control program is also linear, i.e., when the cost, dynamics, and constraints are linear}.  Other risk measures such as the KL divergence-based EVaR metric or the Wasserstein metric provide rich expressions of the uncertainty but are computationally expensive. \addendum{Sampling-based methods for risk computation can also require many samples to guarantee the correctness of the resulting risk-aware plan which remains challenging with computationally-limited hardware, i.e., without a GPU. }

\addendum{\textbf{Risk-Aware Planning with Nonlinear Dynamics.} Keeping with the computation theme, forward computations of nonlinear dynamics with complex policies are already difficult.  Adding risk-aware planning further complicates the issue making such a procedure currently intractable for non-sample-based approaches.  Even still, real-time applications of these methods further limit the number of samples that can be taken to inform planning or control input selection, making efficient estimation of risk measures in these scenarios an open problem.  Recent work indicates that parallelizing computation over a sample-based scenario approach might improve efficiency, but such an approach is still computationally intensive~\cite{akella2023probabilistic}.}

\textbf{Multi-agent interactions.} Our discussion of risk-aware planning and control only considered a single agent. However, real-world dynamic agents may react to the motion of the controlled agent, and these effects could potentially cause unmodeled uncertainty distribution shifts. An important open problem is how to account for the interactions between dynamic agents in a risk-aware manner. 

%\subsubsection{Approximations of risk}
\textbf{Approximations of risk.}  Many risk-aware planning and control techniques either assume that the uncertainty is discrete or use approximation techniques like Sample Average Approximation (SAA) for continuous distributions. How can we guarantee the correctness of risk evaluation and control design when using continuous distribution approximations? The next section introduces methods to verify the risk-aware behavior of autonomous systems. 

\addendum{
\textbf{Constructing Control Barrier Functions}  In the prior risk-aware safety-critical control section, we introduced control barrier functions as a potential method by which a controller can ensure safety even in a risk-aware context.  However, it is worth noting that constructing control barrier functions - let alone the risk-aware formulations - remains an open problem for complex systems.  There has been recent work aimed at learning control barrier functions~\cite{robey2020learning,taylor2020learning,wang2021learning,dai2022learning} and constructing them numerically via methods such as Sum of Squares~\cite{prajna2007convex,clark2021verification,zhang2023efficient} (see~\cite{papachristodoulou2005tutorial} for a tutorial on the sum of squares method).  As such, doing so even in a risk-aware context still remains an open problem.
}
% \newpage

\section{Risk-Aware Verification and Validation}\label{section:VandV}

The previous section provided a high-level summary of risk-aware planning and control, with a more in-depth review of existing literature.  We highlighted the importance of analyzing the inherent uncertainties and risks in robotic operations, particularly when navigating through unstructured environments. This section briefly summarizes the important companion problem of "risk-aware verification and validation" (V\&V) in robotic systems. The verification process determines whether a given system exhibits its desired behavior in the environments in which it is required to operate.  This crucial process ensures that the integrated system performs safely, reliably, and as intended under a broad range of operating conditions.

Integration of risk awareness into the V\&V process allows for a more comprehensive evaluation of a robotic system and its potential to properly respond to risky situations. Specifically, V\&V aims to quantify robotic reliability and safety, explicitly considering interactions with uncertain and unstructured environments~\cite{guiochet2017safety,yoo2009formal,aniculaesei2016towards,grimm2018survey}. As such, risk-aware V\&V requires probabilistic risk assessments, stochastic models, and rigorous testing methods that cover a wide range of potential scenarios. For instance, recent work in Human-Robot Interaction (HRI) has developed procedures to quantify the riskiness of actions taken by systems in a collaborative context~\cite{vicentini2019safety, cheng2022construction, askarpour2016safer, huck2021risk}. These methods typically formalize the risk assessment against existing International Standards, \textit{e.g.} ISO 14121~\cite{ISO_HRI_1} and ISO 12100~\cite{ISO_HRI_2}.  Risk assessment of autonomous systems in other fields has also emerged~\cite{yu2019occlusion,strickland2018deep,ryan2020end,lefevre2014survey}.

As mentioned, prior notions of risk have typically been defined against a corresponding standard and are developed on a case-by-case basis. This observation has prompted recent work in risk-aware verification to adopt the same formal treatment of risk --- tail risk measures --- as used by the synthesis community ~\cite{majumdar2020should,bcbs2013fundamental}. This section delves into the key methodologies and approaches employed for risk-aware verification and validation in robotics, starting with a brief overview of its theoretical foundations and moving toward practical applications and case studies.  As works in this area typically exploit the quantifiable semantics of temporal logics to quantify (un)safe system behavior to make risk-aware verification statements, we begin with a brief overview of temporal logics~\cite{clarke1996formal,woodcock2009formal,nasa_formal}.

\subsection{Temporal Logic}

Temporal logics can be used to express complex system specifications and were originally developed for the analysis and design of reactive systems, i.e., systems with external inputs such as control systems \cite{pnueli1977temporal,baier,clarke1997model}. Temporal logics are extensions of Boolean logic (propositions, negations, conjunctions, disjunctions) by adding temporal operators (until, eventually, always) to reason about the temporal properties of a system. One can make a distinction between temporal logics that reason over qualitative and quantitative temporal properties in their temporal operators. Linear temporal logic, arguably the most common temporal logic, only reasons over qualitative temporal properties, while real-time temporal logics such as metric temporal logic \cite{koymans1990specifying} can reason over quantitative temporal properties. This distinction is best exemplified where a qualitative temporal property is ``eventually reach the goal region'' while its quantitative counterpart could be ``eventually within the next $5$ minutes reach the goal region''. We focus on temporal logic specifications with quantitative temporal reasoning that can encode combinations of timed reachability (``reach region A within $30$ sec''), timed surveillance (``visit regions B, C, and C every $10-60$ sec while agents form a triangular formation"), timed safety (``always between $5-25$ sec stay at least $1$ m away from region E"), and many others.

Temporal logics are formally defined by their syntax and semantics where the syntax defines the rules to construct a temporal logic specification $\phi$ while the semantics define when a temporal logic specification $\phi$ is satisfied (or violated). Spatiotemporal logics, as opposed to temporal logics, also permit reasoning about spatial properties, e.g., allowing a system designer to quantify to which extent (with what safety margin) an obstacle is avoided by a robot.  It is this property that enables us to quantify how well a specification is satisfied \textit{c.f.} how severely a specification is violated, which in turn helps define risk for system verification. Signal temporal logic is a commonly used spatiotemporal logic introduced in \cite{maler2004monitoring}, and we provide a brief introduction to its syntax and semantics in the sidebar \ref{sb:stl_syntax}. For the remainder of the exposition in this section though, assume that $\phi$ denotes a system specification. 

Concerning the use of these logics in verification and validation, over the past decades, the formal methods community proposed and studied a broad range of system verification techniques.  Existing techniques focus on the verification of 1) deterministic systems, or 2) uncertain systems with the two previously discussed viewpoints of the risk-neutral and the worst-case paradigms. In fact, automated verification tools were developed for deterministic systems, e.g.,  model checking \cite{baier,clarke1997model} or theorem proving \cite{shoukry2017smc,sheeran2000checking}. Verification of uncertain systems was particularly studied in the risk-neutral paradigm using probabilistic model checking \cite{bianco1995model,hansson1994logic,kwiatkowska2011prism}, an extension of deterministic model checking, or statistical model checking \cite{younes2002probabilistic,younes2006statistical,legay2019,legay2010statistical}, which are sampling-based techniques for probabilistic system verification.  System verification techniques, however, should not only be able to reason about the probability of violating a specification but also be able to reason about the severity of a violation in terms of rare harmful outcomes. As briefly discussed previously, spatiotemporal logics enable us to quantify how well (severely) a specification is satisfied (violated), and in turn, allow us to define risk in terms of this quantitative measure for stochastic systems.

\begin{open_prob}[Signal Temporal Logic: Syntax and Semantics]\label{sb:stl_syntax}

% \section[Signal Temporal Logic]{Signal Temporal Logic: Syntax and Semantics}
Signal temporal logic (STL) specifications are interpreted over continuous-time signals $\boldsymbol{x}:\mathbb{R}_{\ge 0}\to \mathbb{R}^n$. An STL specification $\phi$ is recursively constructed from atomic predicates by using Boolean operators and temporal operators. These atomic predicates are Boolean-valued functions $\mu:\mathbb{R}^n\to\{1,0\}$ whose truth value is obtained after evaluation of a real-valued function $b:\mathbb{R}^n\to\mathbb{R}$. At time $t$, we obtain the truth value of $\mu$ as
\begin{align}\label{eq:STL_predicate}
\mu({x}(t))\coloneqq\begin{cases}
1 & \text{if } b({x}(t))\ge 0\\
0 &\text{otherwise.}
\end{cases}
\end{align}
Predicate functions $h$ can encode relationships between state variables, such as relative or absolute distances. The syntax of STL  defines a set of rules according to which well-defined STL specifications can be constructed and is given as
\begin{align}\label{eq:full_STL}
\phi \; ::= \; 1 \; | \; \mu \; | \;  \neg \phi' \; | \; \phi' \wedge \phi'' \; | \; \phi'  U_I \phi''
\end{align}
where the operators $\neg$, $\wedge$, and $U_I$ encode negations, conjunctions, and the until over the time interval $I\subseteq \mathbb{R}_{\ge 0}$, respectively. The syntax in \eqref{eq:full_STL} can be understood as follows: the symbol $::=$ assigns one of the expressions from the right-hand side, which are separated by vertical bars, to the free variable $\phi$ on the left-hand side. The variables $\phi'$ and $\phi''$ on the right-hand side are already well-defined STL specifications. While the meaning of negations and conjunctions is clear, the until operator $\phi'  U_I \phi''$ encodes that $\phi'$ has to hold until $\phi''$ holds, which has to happen within the time interval $I$. We can now use logical equivalences to derive the Boolean disjunction, implication, and equivalence operators and the temporal eventually and always operators.  In what follows, $\top$ denotes True in the corresponding logical specification:

\begin{align*}
\phi' \vee \phi''&\coloneqq\neg(\neg\phi' \wedge \neg\phi'') &\text{ (disjunction)},\\
    \phi'\Rightarrow \phi'' &\coloneqq \neg \phi' \vee \phi'' &\text{ (implication)},\\
    \phi'\Leftrightarrow \phi'' &\coloneqq (\phi'\Rightarrow\phi'')\wedge(\phi''\Rightarrow\phi') &\text{ (equivalence)},\\
F_I\phi'&\coloneqq\top U_I \phi' &\text{ (eventually)},\\
G_I\phi&\coloneqq\neg F_I\neg \phi' &\text{ (always)}.
\end{align*}

The semantics of STL define when a signal $x$ satisfies an STL specification $\phi$. These semantics are formally defined as a relation $\models$ between $x$ and $\phi$, and $(x, t)\models \phi$ means that the signal $x$ satisfies the specification $\phi$
at time $t$. We recursively define the semantics as
	\begin{align*}
	 (x,t) \models 1  &\;\;\;\text{ iff } \;\;\;	\text{holds by definition}, \\
	 ({x},t) \models
	 \mu   &\;\;\;\text{ iff }\;\;\;	h({x}(t))\ge 0\\
	 ({x},t) \models \neg\phi' &\;\;\;\text{ iff }\;\;\; ({x},t) \not\models \phi'\\
	 ({x},t) \models \phi' \wedge \phi'' &\;\;\;\text{ iff }\;\;\; ({x},t) \models \phi' \text{ and } ({x},t) \models \phi''\\
	 ({x},t) \models \phi' U_I \phi'' &\;\;\;\text{ iff }\;\;\; \exists t'' \in t\oplus I \text{ s.t. }({x},t'')\models \phi'' \text{ and } \\& \qquad \quad \forall t'\in (t,t'') \text{, }({x},t') \models \phi'.
	\end{align*}
\newpage
\end{open_prob}

\subsection{Motivations for Tail Risk Measures in Verification}
As mentioned previously then, the existence of these spatiotemporal logics makes tail risk measures uniquely suited to serve as the risk measure of choice for verification and validation, and this section will provide an example supporting that claim using the notation offered in Sidebar~\ref{sb:stl_robustness}.  Consider for the sake of argument that we have two controlled systems $\Sigma_1,\Sigma_2$, a Signal Temporal Logic specification $\phi$ denoting the desired behavior required of both systems and a robustness measure $\rho^{\phi}$ for the same specification $\phi$.  For context, every signal temporal logic specification $\phi$ comes equipped with a quantitative measure.  Let's further assume that every time we query either system $\Sigma_1$ or $\Sigma_2$, we receive a random trajectory $\trajectory_1$ or $\trajectory_2$ respectively.  Let $R_i$, $i=1,2$, be a random variable whose samples $r_i$ are the robustness of the trajectories sampled from system $\Sigma_i$, \textit{i.e.} $r_i = \rho^{\phi}(\trajectory_i)$.  Now, let's assume that an oracle tells us that for both systems, with probability $1-\beta$ for some $\beta \in (0,1]$, the random robustness exceeds a cutoff value $\epsilon > 0$.  Furthermore, with probability $\beta$, $\Sigma_1$ exhibits robustness values between $\epsilon$ and $0$, whereas $\Sigma_2$ exhibits robustness values strictly less than $0$.  In other words, $\Sigma_1$ will always realize the desired behavior, while $\Sigma_2$ will, in some rare cases, be unable to realize the desired behavior.  

Put into the context of tail risk measures, both systems exhibit a robustness value-at-risk at risk-level $\beta$ that is positive.  This fact arises as the oracle mentioned that with minimum probability $1-\beta$, both systems exhibit robustnesses exceeding $\epsilon > 0$, \textit{i.e.} $\var_{\beta}(R_i) = \epsilon > 0$. Ending the analysis here results in the correct conclusion that both systems exhibit some minimum probability of realizing satisfactory behavior, and this is a traditional, probabilistic V\&V statement.  However, by considering conditional value-at-risk, we can further discriminate between the two systems, as system $1$ is expected to exhibit satisfactory behavior even in the worst $100\beta \%$ of cases, whereas system $2$ is expected to produce unsatisfactory behavior in the same cases.  This conclusion arises as even in the worse $100 \beta \%$ of cases, the oracle mentioned that system $1$ exhibits robustness values $r_1 \in [0,\epsilon]$ whereas system $2$ exhibits robustness values $r_2 < 0$.  Taking into account the expected value over those cases - the definition of conditional-value-at-risk - we conclude that $\cvar_{\beta}(R_1) \geq 0$ whereas $\cvar_{\beta}(R_2) < 0$.  Therefore, even if both systems exhibit similar minimum probabilities of specification satisfaction, system $1$ is "better" than system $2$ as it is still expected to exhibit satisfactory behavior in the worst $100 \beta \%$ of cases.

The example described above highlights the utility of tail risk measures in risk-aware V\&V.  By considering the robustness value-at-risk, we can make statements on the minimum probability with which a system exhibits a desired behavior in its operating environment(s) - this is the traditional notion of probabilistic V\&V.  Additionally, we can utilize tail risk measures to also make statements on expected worst-case robustness using the conditional-value-at-risk, lower bound such expected worst-case robustness using entropic value-at-risk, and calculate these values without exact distribution knowledge as will be described in sections to follow.

\begin{open_prob}[Signal Temporal Logic: Robust Semantics]\label{sb:stl_robustness}
% \section[Signal Temporal Logic]{Signal Temporal Logic: Robust Semantics}
% \label{sb:stl_robustness}
While the STL semantics tell us if a signal $\boldsymbol{x}:\mathbb{R}_{\geq 0}\to\mathbb{R}^n$ satisfies an STL specification $\phi$, it does not give us any information about the quality of satisfaction. To obtain such information, one can define robust semantics that quantify how robustly the signal $\boldsymbol{x}$ satisfies the specification $\phi$. If $\boldsymbol{x}$ satisfies $\phi$, we would hence like to quantify how different a signal $\boldsymbol{x}^*:\mathbb{R}\to\mathbb{R}^n$ can be from $\boldsymbol{x}$ while still satisfying $\phi$. To quantify this, we first define the closeness of two signals $\boldsymbol{x},\boldsymbol{x}^*:\mathbb{R}\to\mathbb{R}^n$ as
\begin{align*}
d(\boldsymbol{x},\boldsymbol{x}^*)\coloneqq\sup_{t\in\mathbb{R}_{\geq 0}} \|\boldsymbol{x}(t)-\boldsymbol{x}^*(t)\|.
\end{align*}
We now want to compute a value $\rho^\phi$ such that all signals $\boldsymbol{x}^*$ that are such that $d(\boldsymbol{x},\boldsymbol{x}^*)< \rho^\phi$ will also satisfy $\phi$, i.e., $(\boldsymbol{x}^*,t)\models \phi$. To do so, we first define the signed distance of the signal value $\boldsymbol{x}(t)$ to the set of states that satisfy a predicate $\mu$, denoted by $O^\mu:\{x\in\mathbb{R}^n|b(x)\ge 0\}$, as
\begin{align*}
    \text{Dist}(\boldsymbol{x}(t),O^\mu)\coloneqq\begin{cases}
	\inf\limits_{x^*\in \text{cl}(\mathbb{R}^n\setminus  O^\mu) }\|x^*- \boldsymbol{x}(t)\| &\text{ if } \boldsymbol{x} \in O^\mu\\
	- \inf\limits_{x^*\in \text{cl}(O^\mu) }\|x^*- \boldsymbol{x}(t)\| &\text{ otherwise. }
	\end{cases}
\end{align*}
Note that $\text{Dist}(\boldsymbol{x}(t),O^\mu)$ quantifies the extent to which $\mu$ is satisfied (if $\text{Dist}(\boldsymbol{x}(t),O^\mu)>0$) or violated (if $\text{Dist}(\boldsymbol{x}(t),O^\mu)<0$). We can now recursively define the robust semantics of $\phi$ as a real-valued function $\rho^\phi(\boldsymbol{x},t)$ as follows
\begin{align*}
	\rho^{\top}(\boldsymbol{x},t)& \coloneqq \infty,\\
	\rho^{\mu}(\boldsymbol{x},t)& \coloneqq \text{Dist}(\boldsymbol{x}(t),O^\mu),\\
	\rho^{\neg\phi'}(\boldsymbol{x},t) &\coloneqq 	-\rho^{\phi'}(\boldsymbol{x},t),\\
	\rho^{\phi' \wedge \phi''}(\boldsymbol{x},t) &\coloneqq 	\min(\rho^{\phi'}(\boldsymbol{x},t),\rho^{\phi}(\boldsymbol{x},t)),\\
	\rho^{\phi' U_I \phi''}(\boldsymbol{x},t) &\coloneqq \underset{t''\in t\oplus I}{\text{sup}}  \min(\rho^{\phi''}(\boldsymbol{x},t''),\underset{t'\in (t,t'')}{\text{inf}}\rho^{\phi'}(\boldsymbol{x},t') ).
	\end{align*}
Finally, it holds that $(\boldsymbol{x},t)\models \phi$ if and only if $(\boldsymbol{x}^*,t)\models \phi$ for all signals $\boldsymbol{x}^*:\mathbb{R}\to\mathbb{R}^n$ that are such that $d(\boldsymbol{x},\boldsymbol{x}^*)<|\rho^\phi(\boldsymbol{x},t)|$.

\end{open_prob}

\subsection{A General Overview of Risk-Aware V\&V}
For most relevant problems, the system to be verified can be recast as a discrete-time system with known state and input spaces and (perhaps) known dynamics and disturbance spaces.  Formally, at some time $t \in \mathbb{Z}_+ = \{0,1,2,\dots\}$, let $\boldsymbol{x}(t) \in \mathcal{X}$ be the system state, $\boldsymbol{u}(t) \in \mathcal{U}$ be the system control input, $\boldsymbol{d}(t) \in \mathcal{D}$ be a randomly sampled disturbance.  Then for some distribution function $\xi: \mathcal{X} \times \mathcal{U} \times \mathbb{Z}_+ \times \mathcal{D} \to [0,1]$ over $\mathcal{D}$,
\begin{subequations}
\label{eq:ver_general_system}
\begin{equation}
    \boldsymbol{x}(t+1) = f(\boldsymbol{x}(t),\boldsymbol{u}(t),\boldsymbol{d}(t)),
\end{equation}
\begin{equation}
 \suchthat \boldsymbol{d}(t) \sim \pi(\boldsymbol{x}(t),\boldsymbol{u}(t),t).
\end{equation}
\end{subequations}

Finally, let $\mathbb{U}: \mathcal{X} \times \Theta \times \mathbb{Z}_+ \to \mathcal{U}$ be the controller closing the loop for the system to be verified.  Note, the controller $\mathbb{U}$ is parameterized with a parameter $\theta \in \Theta$ to account for exogenous, user-specific inputs that may influence controller behavior, \textit{e.g.} parameterized $3$-space locations for packages in a warehouse that a warehouse robot receives on-the-fly from a central command station when a package is required to be collected.

\textbf{Remark.} For systems that operate in adversarial environments or in the presence of obstacles, the disturbance distribution $\xi$ can be defined as the singleton distribution over the adversarial choice at the given state $\boldsymbol{x}(t)$, input $\boldsymbol{u}(t)$, and time $t$.  For more information see dirac distributions~\cite{kanwal1998generalized} and adversarial testing works such as~\cite{akella2020formal,akella2023barrier,zhang2020testing,ghosh2018verifying}.  We can consider both cases --- the case with non-adversarial, randomized disturbances and the case with adversarial or otherwise known disturbances --- in the same stochastic setting. Furthermore, the state and input spaces have been left arbitrary to allow both the continuous space definitions in the controls community and the finite-state and input definitions in the (PO)MDP computer science literature.

Closing the loop between~\eqref{eq:ver_general_system} and the assumed controller $\mathbb{U}$ generates a system $\Sigma$ that when queried with a specific initial condition $x_0 \in \mathcal{X}_0 \subseteq \mathcal{X}$ and controller parameter $\theta \in \Theta$ produces a (perhaps) different state trajectory $\trajectory$, which is an element of the signal space $\signalspace = \{s: \mathbb{Z}_+ \to \mathcal{X}\}$:
\begin{subequations}
    \label{eq:ver_trajectory_sample}
    \begin{equation}
\mathrm{a~sample~of~}\Sigma(x_0,\theta)~\mathrm{is~} \trajectory \triangleq \{x_0 \equiv \boldsymbol{x}(0), \boldsymbol{x}(1), \dots\} 
\end{equation}
\begin{equation}
\suchthat \boldsymbol{x}(t+1) = f(\boldsymbol{x}(t),\mathbb{U}(\boldsymbol{x}(t),\theta),\boldsymbol{d}(t)).        
    \end{equation}
\end{subequations}
Finally, let $\classifier$ be a classifier function mapping from the signal and parameter spaces to the real-line, \textit{i.e.} $\classifier: \signalspace \times \Theta \to [-a,b]$ where parameters $a,b \in \mathbb{R}_{++}$ are finite.  The classifier $\classifier$  delineates between satisfactory behavior --- trajectory and parameter pairs $(\trajectory, \theta)$ that evaluate to a positive value, \textit{i.e.} $\classifier(\trajectory,\theta) \geq 0$ --- and unsatisfactory behavior --- pairs that evaluate to a negative value.  Examples of such a classifier could be the robustness functions $\rho$ from signal temporal logic, the minimum value of a barrier function $h$ over time~\cite{ames2016control}, \textit{etc}.  For a description of robustness measures in Signal Temporal Logic, please see Sidebar~\ref{sb:stl_robustness}. Generally speaking, we define the outcome of function  $\classifier$ to be the \textit{robustness} of the corresponding trajectory and parameter pair, \textit{i.e.} for $r = \classifier(\trajectory,\theta)$, where $r$ is the trajectory and parameter pair's robustness value.  More positive values of $\classifier(\trajectory,\theta)$ indicate better, \textit{more robust} realization of the desired behavior.

\textbf{Remark}  The rationale to analyze multiple, non-unique trajectories $\trajectory$ as defined in~\eqref{eq:ver_trajectory_sample} arises from the fact that disturbances $\boldsymbol{d}$ in~\eqref{eq:ver_general_system} are sampled randomly at each time $t$ from the distribution function $\xi$.  If the distribution function $\xi$ were the singleton distribution corresponding to a specific disturbance $d \in \mathcal{D}$ for each state, input, and time $(x,u,k) \in \mathcal{X} \times \mathcal{U} \times \mathbb{Z}_+$, then $\Sigma(x_0,\theta)$ would always produce the same trajectory $\trajectory$ upon successive queries at $(x_0,\theta)$, and this argument holds $\forall~(x_0,\theta) \in \mathcal{X}_0 \times \Theta$.

\begin{figure*}[t]
    \centering
    \includegraphics[width=\textwidth]{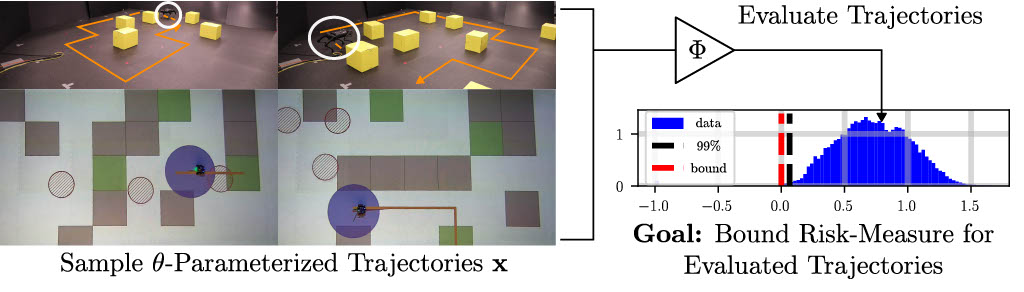}
    \caption{A flowchart for a general risk-aware verification pipeline.  In the figure, the parameters $\theta$ correspond to obstacle locations and waypoints for the robots highlighted by white and blue circles~\cite{akella2022safety}.  Risk-aware verification bounds the risk-measure evaluation of evaluated trajectories --- the blue data shown in the right figure.}
    \label{fig:verification_flowchart}
\end{figure*}

\addendum{The goal of risk-aware verification then, is to bound the risk-measure evaluation of the classifications of these (perhaps) stochastically evolving trajectories.  Stated formally, let $\chi$ be a tail risk measure, \textit{e.g.} Value-at-Risk, Conditional-Value-at-Risk, Entropic-Value-at-Risk, \textit{etc}.  Then, at some risk-level $\beta \in [0,1]$, determine an upper or lower bound to $\chi\left(\classifier\left(\Sigma(x_0,\theta)\right)\right)$ for some $(x_0,\theta) \in \mathcal{X}_0 \times \Theta$.  Figure~\ref{fig:verification_flowchart} depicts this generic risk-aware verification pipeline, and the following section will specify how this pipeline has been implemented in a variety of recent works.  To facilitate that discussion, we will define the \textit{robustness} $R(x_0,\theta)$ to be the random variable whose samples $r = \classifier(\trajectory, \theta)$, where $\trajectory$ is a sample of the random variable $\Sigma(x_0,\theta)$.  In other words, $R(x_0,\theta) = \classifier(\Sigma(x_0,\theta),\theta)$ --- this term was first defined in~\cite{akella2022sample}.} 

\subsubsection{Examples}
Perhaps the most prevalent examples of risk-aware verification arise from a re-framing of traditional work in the Stochastic Model Checking (SMC) community~\cite{kwiatkowska2007stochastic,sen2005statistical,legay2010statistical,agha2018survey}.  With respect to the aforementioned pipeline, SMC assumes the ability to collect system traces --- trajectories $\trajectory$
--- and evaluate their satisfaction of a desired behavior.  These behaviors are typically expressed as a specification $\phi$ in Probabilistic Computational Tree Logic ~\cite{ciesinski2004probabilistic}, which is a form of Temporal Logic (see Sidebar \ref{sb:stl_syntax}).  As such, each of these behaviors has satisfiability metrics --- classifier functions $\classifier$ in our overarching methodology --- with which to determine trace satisfaction of the desired behavior $\phi$.  

SMC consists of two different analyses.  The first, hypothesis testing, asserts that the system $\Sigma$ realizes the behavior $\phi$ with minimum probability $p$ and determines the minimum number of system traces that have to be evaluated to accept or reject this hypothesis.  The second, estimation, exploits either the Chernoff bound or Hoeffding's inequality to estimate the probability $p$ with which $\Sigma$ realizes $\phi$ within some tolerance bounds that are a function of the number of trajectories sampled and evaluated.  In both cases, however, the probability of satisfaction $p$ has a one-to-one correspondence with the Value-at-Risk of the random variable $R(x_0)$ (we omit $\theta$ in the notation here, as SMC typically does not consider parameterized trajectories).  More specifically, $p$ is such that $\var_{1-p}(R(x_0)) \geq 0$.

These are not the only works that take a Value-at-Risk approach to system verification.  In~\cite{akella2022scenario}, the authors use scenario optimization to lower bound the Value-at-Risk of the robustness random variable $R(x_0,\theta)$ for a user-defined $\beta \in [0,1]$.  Similarly, the authors of~\cite{lindemann2021stl} use a sample-average-approximation procedure to estimate the Value-at-Risk of the same robustness variable for any user-defined $\beta \in [0,1]$.  In~\cite{cubuktepe2018verification}, the authors go one step further and express policy or controller synthesis as an optimization problem over a general class of risk measures for verification purposes.  They show numerical examples of the success of a convex-concave procedure in identifying such policies for a Markov Decision Process.  In this case, the policies optimize for a certain risk sensitivity as expressed by Value-at-Risk among other risk measures expressed in Cumulative Prospect Theory~\cite{wakker1993axiomatization}.  Finally, in~\cite{leung2022semi}, the authors modify a learned controller online whenever the learned controller outputs an infeasible trajectory.  Via a gradient-descent method, they update controller parameters until the resulting trajectory passes an intermediary risk-aware verification step, before implementation of the modified controller.  In general, however, any of the aforementioned risk-aware works could also be conceived of as Value-at-Risk-based verification, insofar as the classifier functions $\classifier$ were developed against specific standards for their respective applications~\cite{vicentini2019safety, cheng2022construction, askarpour2016safer, huck2021risk, yu2019occlusion,strickland2018deep,ryan2020end,lefevre2014survey}

\begin{figure*}[t]
\centering
\includegraphics[width = \columnwidth]{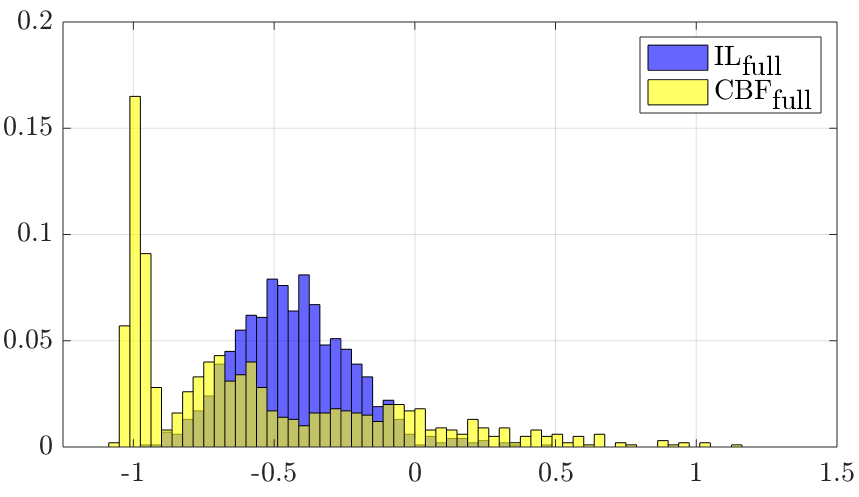}
\includegraphics[width = \columnwidth]{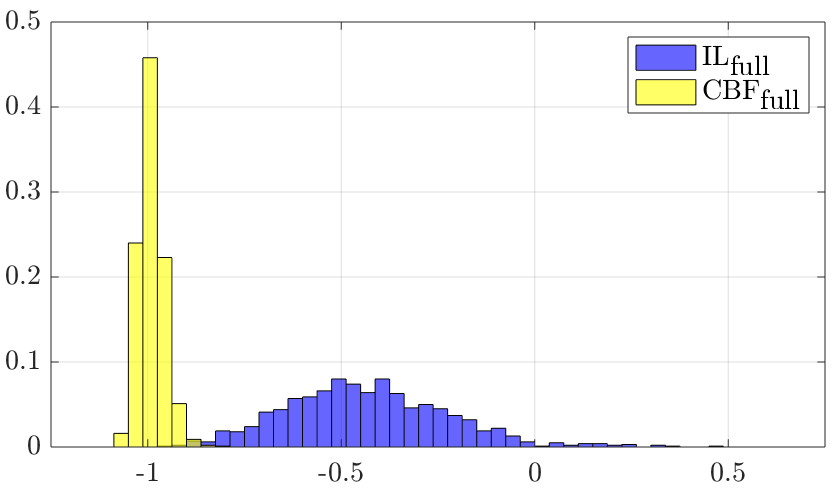}
\includegraphics[width = \columnwidth]{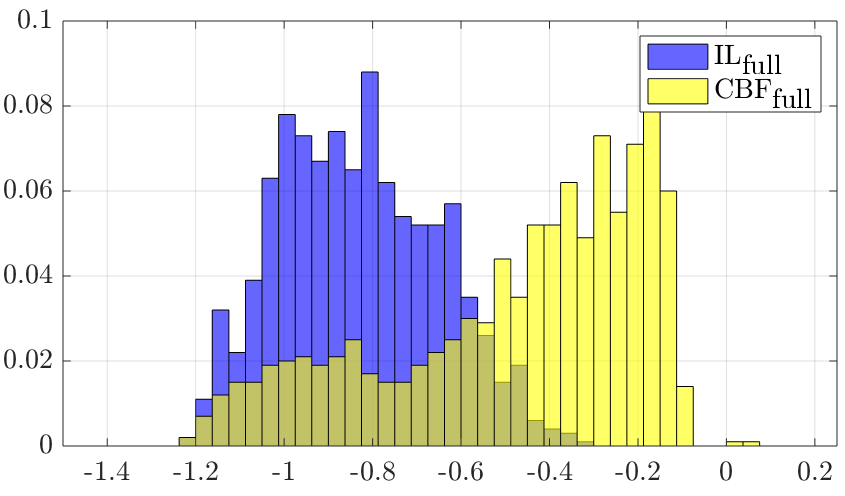}
\includegraphics[width = \columnwidth]{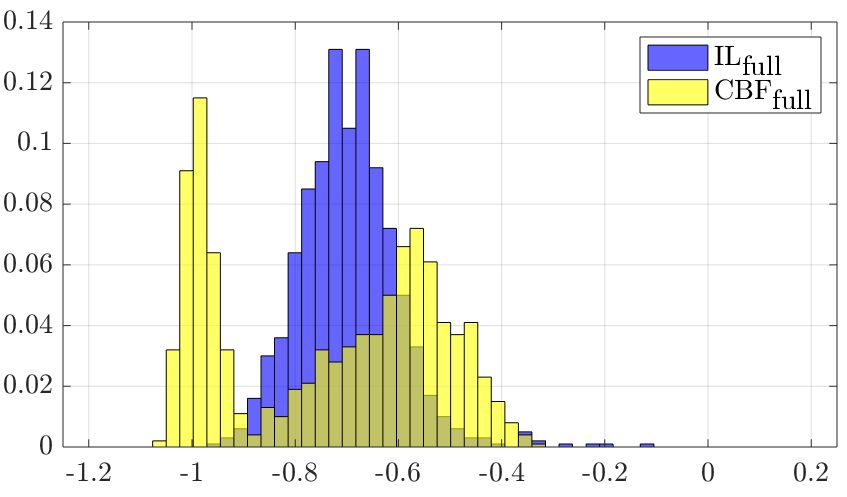}
\caption{Empirical distributions for the Imitation Learning (IL) and Control Barrier Function-based (CBF) controllers and for the specifications $\phi_1$-$\phi_4$ as described in the case study on risk-aware lane-keeping. Figures taken from \cite{lindemann2022risk}.}
\label{fig:CARLA__}
\end{figure*}

\begin{figure}[t]
\centering
\includegraphics[width = \columnwidth]{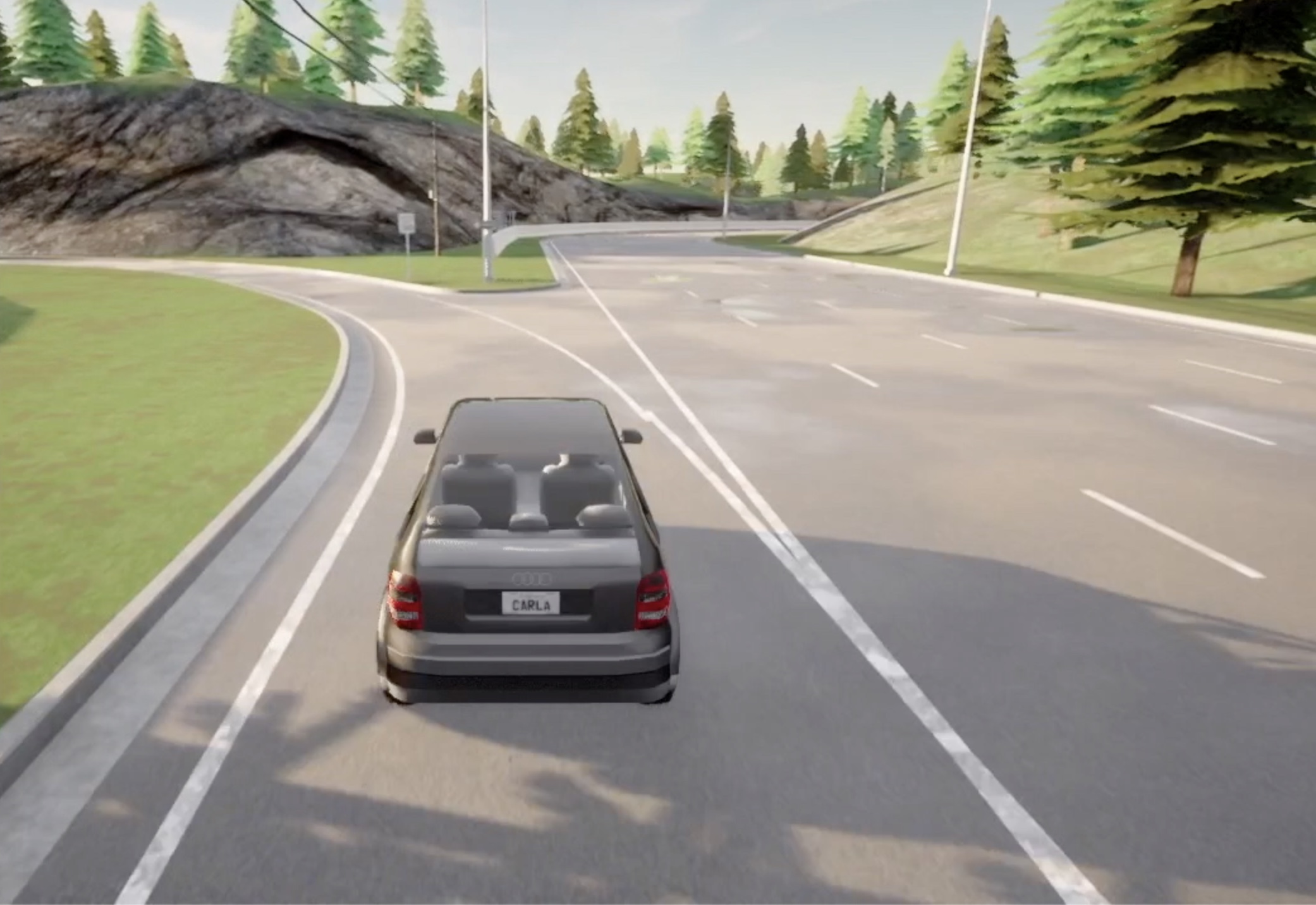}
\includegraphics[width = \columnwidth]{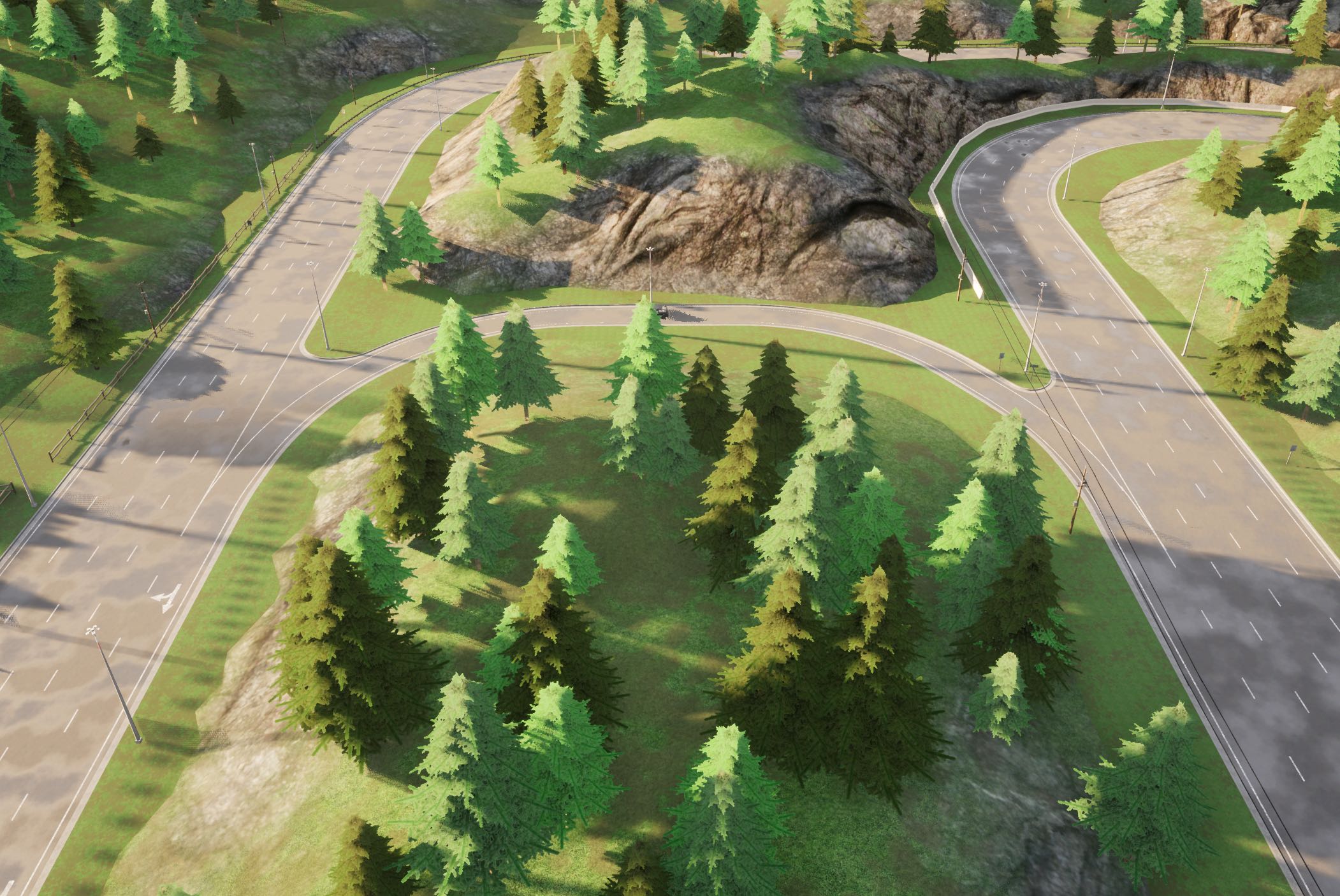}\vspace{0.1 in} \\ 
\includegraphics[scale=0.365]{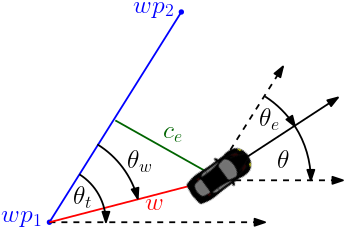} \vspace{0.1 in} \\
\caption{Top: Simulation environment in the CARLA autonomous driving simulator. Middle: Left turn on which we evaluate two neural network lane-keeping controllers. Bottom: The car's cross-track error $c_e$ and orientation error $\theta_e$  with respect to waypoints. Figures taken from \cite{lindemann2022risk}.}
\label{fig:CARLA_}
\end{figure}

However, Value-at-Risk verification represents a smaller fraction of risk-aware verification efforts as compared to works using coherent risk measures, such as Conditional-Value-at-Risk.  For example, in a similar paradigm as in~\cite{cubuktepe2018verification}, in~\cite{meggendorfer2021verification} the author proves that there exist polynomial time algorithms to determine policies for an MDP that are verifiable by default.  Verification arises as the policies are synthesized to achieve a minimum conditional value at risk with respect to objective satisfaction.  Similarly, in~\cite{samuelson2018safety} the authors utilize a CVaR constraint for their optimal controller and verify that the system remains within a risk-sensitive safe set defined by the same CVaR constraint.  In~\cite{hashemi2022risk}, the authors develop a procedure for learning a controller to tackle simultaneous performance and safety tradeoffs for nonlinear systems and verify the learned controller by estimating the CVaR of a corresponding robustness random variable.  In~\cite{quagliarella2017robust}, the authors constrain the optimal design of supersonic aircraft bodies against both VaR and CVaR requirements to ensure verifiable, risk-aware performance despite uncertainties arising from the transition from laminar to turbulent flow, manufacturing uncertainties, \textit{etc}. Examples from a larger body of work, including those by the authors, can be found in references~\cite{chapman2021risk,lindemann2021stl,lindemann2022risk,lindemann2022temporal,akella2022sample,rigter2021risk,godbout2021carl,hong2011monte,kim2015conditional,kim2022efficient, samuelson2018safety, chapman2022optimizing, weifausschapmanconf2022}.

\subsubsection{Case Study: Risk-Aware Verification of Lane Keeping Controllers}
In this case study, the goal is to find the least risky controller among two neural network lane-keeping controllers in the autonomous driving simulator CARLA during a left-turn \cite{dosovitskiy2017carla}, see  Fig. \ref{fig:CARLA_} (top and middle). Lanekeeping is realized by tracking a set of predefined waypoints. For verification, we consider the cross-track error $c_e$ and the orientation error $\theta_e$ with respect to the current and next waypoint, see Fig. \ref{fig:CARLA_} (bottom). We consider an imitation learning (IL) controller~\cite{ross2010efficient} and a control barrier function (CBF) controller~\cite{lindemann2021learning}. The car model is stochastic, as the control inputs are subject to normally distributed noise, and we uniformly sample the car's initial position from the set  $(c_e,\theta_e)\in [-1,1]\times [-0.4,0.4]$. We collected $N\coloneqq1000$ trajectories $\trajectory_i$ in a validation set $D_\text{val}$ for each controller. 
% From a visual inspection, we can not immediately assess which controller is more risky. 

We are first concerned with cross-track error and consider the specifications $\phi_1\coloneqq G_{[0,\infty)} (|c_e|\le 2.25)$ where $2.25$ is an empirically obtained threshold indicating that the car stays within the lane. For the following analysis, recall that a negative value of $-\rho^{\phi_1}$ indicates satisfaction of $\phi_1$ and positive values indicate a failure to lane-keep. Upper bounds on  ${VaR}_{0.95}(-\rho^{\phi_1}(\trajectory))$, ${CVaR}_{0.85}(-\rho^{\phi_1}(\boldsymbol{x}))$, and ${E}(-\rho^{\phi_1}(\boldsymbol{x}))$ are reported in Table~\ref{tab:lane_keeping_data}, along with the empirical satisfaction rate $P_{\phi_1}\coloneqq|\{\trajectory_i\in D_\text{val}|\boldsymbol{x}_i \text{ satisfies } \phi_1\}|/N$.
\begin{table}
    \centering
    \begin{tabular}{|p{0.35\columnwidth}|p{0.2\columnwidth}|p{0.2\columnwidth}|}
    \hline
    \backslashbox{$R$}{~~Controller} & IL & CBF \\
    \hline
    $\overline{VaR}_{0.95}(-\rho^{\phi_1}(\boldsymbol{x}))$ & \textcolor{green}{0.462} & 1.125\\
    $\overline{CVaR}_{0.85}(-\rho^{\phi_1}(\boldsymbol{x}))$ & \textcolor{green}{1.436} & 1.818 \\
    $\overline{E}(-\rho^{\phi_1}(\boldsymbol{x}))$ & -0.248 & \textcolor{green}{-0.375} \\
    $P_{\phi_1}$ & \textcolor{green}{0.975} & 0.863 \\
    $\overline{VaR}_{0.95}(-\rho^{\phi_2}(\boldsymbol{x}))$ & 0.462 & \textcolor{green}{-0.794} \\
    $\overline{E}(-\rho^{\phi_2}(\boldsymbol{x}))$ & -0.254 & \textcolor{green}{-0.81} \\
    $\overline{VaR}_{0.95}(-\rho^{\phi_3}(\boldsymbol{x}))$ & \textcolor{green}{-0.324} & 0.063 \\
    $\overline{E}(-\rho^{\phi_3}(\boldsymbol{x}))$ & \textcolor{green}{-0.652} & -0.297 \\
    $\overline{VaR}_{0.95}(-\rho^{\phi_4}(\boldsymbol{x}))$ & -0.13 & \textcolor{green}{-0.32} \\
    $\overline{E}(-\rho^{\phi_4}(\boldsymbol{x}))$ & -0.517 & \textcolor{green}{-0.533} \\
    $P_{\phi_5}$ & 1 & 1 \\
    \hline
    \end{tabular}
    \caption{Tabulated data from~\cite{lindemann2022risk} for the case study on risk-aware lane-keeping of the Imitation Learning (IL) and Control Barrier Function (CBF) controllers.}
    \label{tab:lane_keeping_data}
\end{table}
Clearly, the IL controller is the least risky one in terms of $VaR_{0.95}$ and $CVaR_{0.85}$, and it also has the highest empirical satisfaction rate. Interestingly though, the CBF controller performs better on average. This result can also be seen in the empirical histograms of Fig. \ref{fig:CARLA__} (top left). We hypothesize that this behavior arises from the long tail of risky behavior for the CBF controller, which corresponds to transient system behavior.  We also analyzed the controllers' behavior more closely by looking at the cross-track error during the steady-state and transient phases for the specifications $\phi_2\coloneqq G_{[10,\infty)} (|c_e|\le 2.25)$ and $\phi_3\coloneqq F_{[0,5]}G_{[0,5]}(|c_e|\le 1.25)$, respectively. The upper bounds of the  ${VaR}_{0.95}(-\rho^{\phi_i}(\boldsymbol{x}))$ and ${E}(-\rho^{\phi_i}(\boldsymbol{x}))$ for $\phi_2$ and $\phi_2$ are shown in Table~\ref{tab:lane_keeping_data} as well.

Interestingly, the IL controller is the least risky one only during the transient phase, while the CBF controller is the least risky one in steady state. The corresponding empirical distributions are shown in  Figs. \ref{fig:CARLA__} (top right and bottom left). Finally, let us verify the controller risk in terms of the orientation error $\theta_e$. Consider $\phi_4\coloneqq G_{[0,\infty)} \big((c_e\ge 1.25) \implies F_{[0,2]}G_{[0,1]}(\theta_e\le 0)\wedge (c_e\le -1.25) \implies F_{[0,2]}G_{[0,1]}(\theta_e\ge 0)\big)$ which expresses the need for the controller to react to large cross-track errors $c_e$ using the right orientation adjustment. 

For this specification, the CBF controller is the least risky controller, which aligns with our observation that it is a better controller during steady-state. It can further be observed that both controllers have the same empirical satisfaction probability, while our risk analysis better quantifies a controller's quality. The empirical distribution of both controllers is shown in Fig. \ref{fig:CARLA__} (bottom right).

\subsubsection{Case Study: Risk-Aware Verification of Quadrupedal Locomotion}
{\begin{figure}
    \includegraphics[width = \columnwidth]{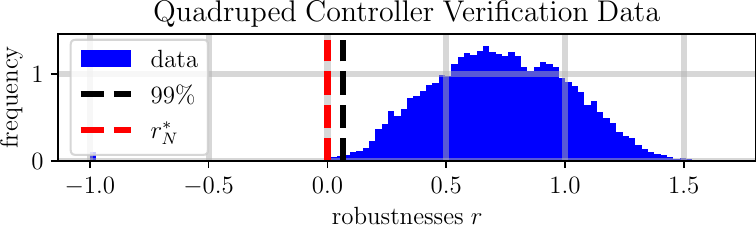}
    \includegraphics[width = \columnwidth]
    {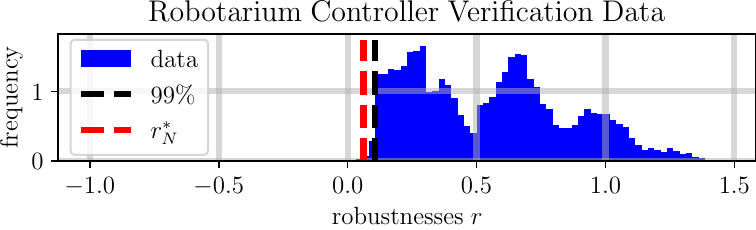}
    \captionof{figure}{Validation data for probabilistic lower bounds reported on $\var_{0.99}(R)$ for controllers generated for a quadruped (top) and robotarium (bottom).  As shown, the reported lower bounds (red) generated by the scenario approach mentioned in Sidebar~\ref{sb:concentration_inequalities} are accurate as they lower bounds the true $\var_{0.99}(R)$ (black).  Information for this figure comes from~\cite{akella2022barrier}.}
    \label{fig:case_study_validation_data}
\end{figure}
}

This case study aims to verify a quadruped's ability to render positive a collision-avoidance barrier function $h$ for at least $T = 150$ time-steps with a time-step $\Delta t = 0.1$ ~\cite{akella2022barrier}.  Keeping consistent with our notation for the general overview for risk-aware V\&V, our parameters $\theta$ include the locations of $4$ randomly placed static obstacles in a $5 \times 5$ meter grid, and the center coordinates of a goal region $g$ in the same grid.  Hence, $\theta \in [-5,5]^{10} \triangleq \Theta$.  To simplify our analysis, we represent the quadruped as a unicycle system, and as such, we assume we can initialize the quadruped at a random planar position and angular orientation in the grid, \textit{i.e.} $x_0 \in [-5,5]^2 \times [0,2\pi]\triangleq \mathcal{X}_0$.  Our classifier function $\classifier$ evaluates the discrete-time fractional difference of a candidate barrier function $h$ that the quadruped is to keep positive.  As such, the classifier outputs the minimum value over all time-steps $k$ of $h(\boldsymbol{x}(t+1)/h(\boldsymbol{x}(t))$, as realized by the quadruped over one trajectory $\trajectory = \{\boldsymbol{x}(0),\boldsymbol{x}(1),\dots,\boldsymbol{x}(150)\}$.  Therefore, $\classifier(\trajectory) < 0$ is equivalent to stating that there existed an interior time-step $x_j \in \trajectory$ such that $h(x_j) < 0$ and the quadruped failed to remain safe.

\begin{figure}
    \centering
    \includegraphics[width = 0.49\columnwidth]{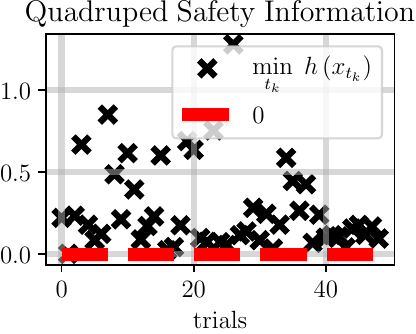}
    \includegraphics[width = 0.49\columnwidth]{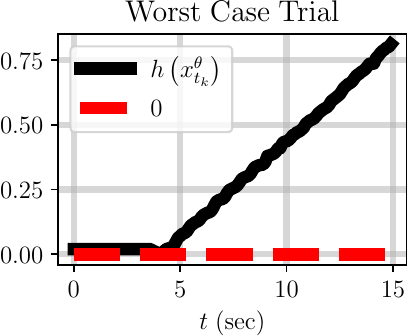}
    \caption{Worst case safety information for the quadrupedal case study.  Over the $50$ sampled trials, the quadruped realizes a positive barrier value every time, which, according to the concentration inequality results in Sidebar~\ref{sb:concentration_inequalities}, implies that the system should always keep positive the barrier function $h$ with minimum probability $90\%$ and with confidence $99.5\%$.  Information for this figure comes from~\cite{akella2022barrier}.}
    \label{fig:quad_case_study_info}
\end{figure}

Slightly different from the general overview, however, instead of aiming to determine a lower bound on the Value-at-Risk level $\beta = 0.9$ of the robustness random variable $R(x_0,\theta)$, we also randomize over initial conditions and parameters $(x_0,\theta)$ from their combined space.  As such, the evaluation $r$ of a sampled trajectory $\trajectory$ generated by first sampling $(x_0,\theta)\sim\uniform[\mathcal{X}_0 \times \Theta]$ is a sample of the \textit{holistic robustness random variable} $R$ --- this term was first defined in~\cite{akella2022sample}.  That being said, we still aim to lower bound $\var_{\beta = 0.9}(R)$, which, according to the sample-based methods detailed in Sidebar~\ref{sb:concentration_inequalities}, has a known sample complexity (number $N$ of trajectories to be evaluated) to determine such a lower bound.  \addendum{To be specific, risk-aware verification in this context amounts to identifying a high-probability lower bound on the Value-at-Risk of the holistic robustness random variable $R$ through a sample-based approach predicated on scenario optimization~\cite{campi2008exactfeasibility}.}  Therefore, after taking $N=50$ trajectories, we can state with $\approx 99.5\%$ confidence, that the quadruped will realize positive value trajectories with $90\%$ probability, as the identified lower bound on $\var_{\beta = 0.9}(R)$ was positive.  The associated safety information is depicted in Figure~\ref{fig:quad_case_study_info}.

We can also show that the reported probabilistic bounds are accurate.  In~\cite{akella2022safety} we employed the same, risk-aware verification procedure as described above, to validate the controllers for a Quadruped and a multi-agent robotic system~\cite{wilson2020robotarium}.  The expressions for the classifier function were the same in both cases, though we will refrain from reproducing them here for the sake of brevity.  Suffice it to say that any trajectory that evaluates to a positive value under $\classifier$ would have made non-trivial progress toward a goal while avoiding static/moving obstacles within $10$ seconds.  To that end then, we only implemented controllers on hardware systems once they exhibited a positive lower bound for $\var_{0.99}(R)$.  To determine such a lower bound we sampled $300$ trajectories for both controllers and calculated their robustness under the aforementioned classifier $\classifier$.  Doing so for our chosen controllers indicated positive lower bounds, and we can verify the accuracy of these lower bounds by taking $20000$ trajectories, evaluating them, numerically approximating the distribution of the holistic robustness random variable $R$, and reporting the numeric $\var_{0.99}(R)$ as our approximation.  Figure~\ref{fig:case_study_validation_data} showcases the validity of the reported lower bounds, overlaid on the numeric approximation of the distribution for $R$, and Figure~\ref{fig:case_study_validated_controllers} shows tiles of the controllers implemented on their respective hardware systems.  Note that in all of the randomized cases, the controllers steered the systems successfully to their goals while avoiding all obstacles.  This controller reliability is the primary reason we take a tail risk approach to verification, as the purpose of the procedure is to identify rare, unsafe phenomena and ensure that even in those rare cases, the system still performs admirably. 

\begin{open_prob}[Concentration Inequalities: Risk Measure Estimation]\label{sb:concentration_inequalities}
This sidebar will briefly describe two methods to estimate the tail risk measures expounded upon in this article.  We will describe these methods as they are applied to arbitrary scalar random variables $X$.

\textbf{Sample-Average Approximation} This first method estimates $\var_{\beta}(X),~\cvar_{\beta}(X)$ for any $\beta \in (0,1)$ and any scalar random variable $X$.  Let $\{x_i\}_{i=1}^N$ be a set of $N$ independently drawn samples of $X$.  The empirical distribution function $\empiricaldist(x)$ for $X$ based on this set of samples is

\begin{equation}
    \empiricaldist(x) = \frac{1}{N}\sum_{i=1}^N~\indicator(x \leq x_i),~\forall~x_i \in \{x_i\}_{i=1}^N.
\end{equation}
with $\indicator$ being the indicator function.  The Sample-Average Approximation (SAA) exploits the Dvoretsky-Kiefer-Wolfowitz Inequality~\cite{dvoretzky1956minimax} built upon by Paul Massart in~\cite{massart1990DKWtight}, which proves that the empirical distribution has bounded deviation with respect to the true cumulative distribution function $F$ for $X$ to within some probability $\delta \in (0,1)$:
\begin{subequations}
    \label{ci:eq:dvoretsky_ineq}
\begin{align}
  &  \empiricaldist(x)  - \sqrt{\frac{1}{2N}\ln\left(\frac{2}{\delta}\right)} \leq F(x) \\
   & \leq \empiricaldist(x) + \sqrt{\frac{1}{2N}\ln\left(\frac{2}{\delta}\right)},~\mathrm{w.p.~} \geq 1-\delta.
\end{align}
    
\end{subequations}
The tail risk $\var_{\beta}(X)$ can be lower and upper bounded for any $\beta \in (0,1)$ using (\ref{ci:eq:dvoretsky_ineq}).  Define upper bound $\overline{\var_{\beta}}(X)$ as:
\begin{align*}
    % \label{ci:eq:ub_var}
    &\overline{\var_{\beta}}(X,\delta) = \\ &\inf\left\{x \in \mathbb{R} ~\bigg|~ \empiricaldist(x) - \sqrt{\frac{1}{2N}\ln\left(\frac{2}{\delta}\right)} \geq 1-\beta\right\},
\end{align*}
and let the lower bound $\underline{\var_{\beta}}(X)$ be defined as:
\begin{align*}
    % \label{ci:eq:lb_var}
    &\underline{\var_{\beta}}(X,\delta) = \\ & \inf\left\{x \in \mathbb{R} ~\bigg|~ \empiricaldist(x) + \sqrt{\frac{1}{2N}\ln\left(\frac{2}{\delta}\right)} \geq 1-\beta\right\}.
\end{align*}
Then, using~\eqref{ci:eq:dvoretsky_ineq}, the following result holds~$\forall~\beta,\delta \in (0,1)$
\begin{equation*}
    \underline{\var_{\beta}}(X,\delta) \leq \var_{\beta}(X) \leq \overline{\var_{\beta}}(X,\delta)~\mathrm{w.p.}\geq 1-\delta.
\end{equation*}
Note that as the number of samples, $N$, of the random variable $X$ increases, the gap between the upper and lower bounds shrinks, as the bounds converge to the true value $\var_{\beta}(X)$.
Similar methods exist to estimate $\cvar_{\beta}(X)$ as well~\cite{pmlr-v97-thomas19a}.

\textbf{Scenario Bounds.}  The second method upper bounds $\var_{\beta}(X),\cvar_{\beta}(X),\evar_{\beta}(X)$ for any $\beta \in (0,1)$.  As before, let $\{x_i\}_{i=1}^N$ be a set of $N$ independently drawn samples of the scalar random variable $X$.  Consider the following optimization problem, termed a scenario program~\cite{campi2008exactfeasibility}:
\begin{equation}
    \label{ci:eq:scenario_ub}
    \begin{aligned}
        \zeta^*_N & = \argmin_{\zeta \in \mathbb{R}}~ & &\zeta, \\
        &~~\mathrm{subject~to}~ & & \zeta \geq x_i,~\forall~x_i \in \{x_k\}_{k=1}^N.
    \end{aligned}
\end{equation}
The theory of scenario optimization  states that the solution to this optimization problem is an upper bound on $\var_{\beta}(X)$ with minimum probability $1-(1-\beta)^N$, \textit{i.e.} if $X$ has probability density function $\pi$, then
\begin{equation}
    \label{ci:eq:ub_var}
    \prob_{\pi}^N\left[\zeta^*_N \geq \var_{\beta}(X) \right] \geq 1-(1-\beta)^N.
\end{equation}
The above result was proven in~\cite{akella2022scenario}.  Note that (\ref{ci:eq:ub_var}) does not need the density function $\pi$ for $X$ to be known.  It just requires an ability to take $N$ independent samples of $X$.  Therefore, if we have a constant $c \in \mathbb{R}$ such that $\prob_{\pi}[x \leq c] = 1$, then we can exploit this inequality~\eqref{ci:eq:ub_var} to similarly upper bound $\cvar_{\beta}(X)$ and $\evar_{\beta}(X)$.  Details on this approach can be found in Section~3 of~\cite{akella2022sample}.

\addendum{\textbf{Remark on Utility in Controller Synthesis and Verification:} As will be mentioned in various sections in this article, utilizing these concentration inequalities to estimate risk measures assumes that any taken samples are identically distributed.  This may pose a problem for controller synthesis, as oftentimes the randomness prompting the need for a risk-aware analysis stems from a random disturbance whose distribution is conditioned on the control input to be chosen.  Similarly for verification, one must take care to ensure that the method used to randomly sample trajectories to be verified consistently samples these random trajectories from the same distribution.  As ensuring that these samples are identically distributed oftentimes requires considerable effort or may be impossible, recent work aims to estimate these risk measures over non-stationary distributions, though this remains an open problem (for more here, please reference the Open Problems and Future Directions section towards the end of the article).}
\end{open_prob}

{
\centering
\includegraphics[width = \columnwidth]{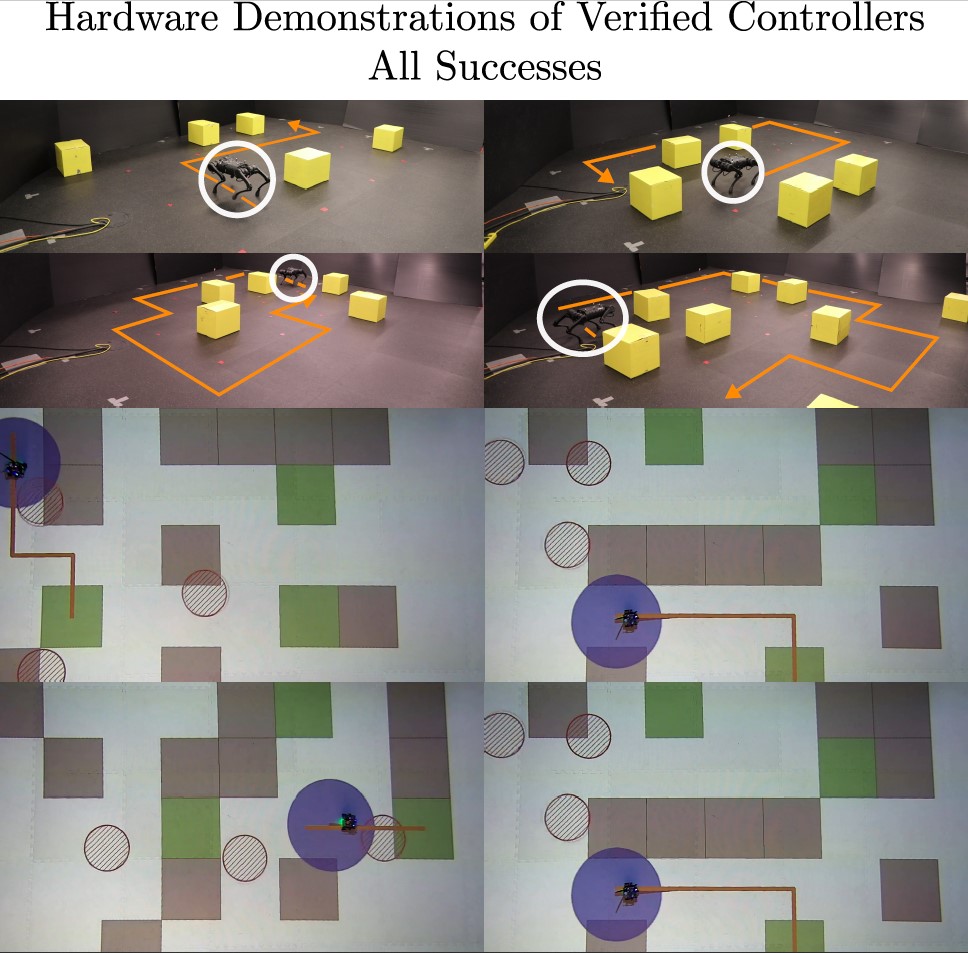}
\captionof{figure}{Hardware demonstration of controllers verified from a tail risk perspective.  As the implemented controllers were ``verified" since the reported lower bound on their Robustness Value-at-Risk was positive, we expect decent behavior in practice.  Indeed, the verified controllers performed admirably despite randomized test cases generated for each system.  Figure adapted from~\cite{akella2022safety}.}
\label{fig:case_study_validated_controllers}
}

\subsection{Open Questions and Future Directions in V\&V}
\subsubsection{Sample Complexity} There exist several open questions in risk-aware verification, though the most notable one concerns the small tail probability requirements for industrial applications.  More specifically, for most product-level robotic systems requiring a verification statement, \textit{i.e.} autonomous cars, factory robots, flight software, \textit{etc}, current standards require these systems to be verified to arbitrarily high probabilities, \textit{i.e.} $1-10^{-4}, 1-10^{-6}, 1-10^{-9}$ or even higher.  If we assume that the underlying distributions are known, \textit{i.e.} Gaussian as is typically done, then this analysis can be carried out in a tractable, even analytic, fashion.  However, if we follow the philosophy underlying the sample-based works that have recently become more popular, as they do not assume underlying distributional knowledge for verification, then verifying systems to these probabilities could require hundreds of thousands of samples or even more.  If each of those samples constitutes even one experimental run of the system, then this makes the direct application of these theoretical concepts exceedingly costly or time-intensive.  As such, reducing this sample complexity, whether via intelligent test design or by leveraging partial system knowledge, would go a long way to facilitate the widespread industrial adoption of these currently theoretical pipelines.

\subsubsection{Compositional Verification} In a similar vein as prior, this second open question stems from a primarily industrial concern as well.  Namely, typical large-scale systems are composed of a variety of moving parts, each of which has to satisfy its own component specifications such that the larger, architectured system satisfies a grander objective.  Per the prior pipeline, each component could be verified separately in a probabilistic fashion, but in systems with potentially hundreds of separately engineered parts, separate verification procedures could be potentially prohibitive.  On the other hand, the system could be verified as a large-scale black-box system, though this could similarly fall under the sample-complexity questions as risen in the prior subsection.  As such, determining an optimal way of breaking down these larger-scale, system-level specifications, into easily verifiable subcomponents for their respective systems remains an open problem.  Indeed, the satisfiability of a given signal temporal logic specification itself remains a challenging problem.  Determining the minimum number of such subcomponents would also mitigate any further sample-complexity issues arising from separate verification procedures as well.  On the other hand, perhaps via smart instrumentation, all verification procedures for all subcomponents could be performed simultaneously.

\section{Open Problems and Future Directions}

Our discourse up to this point has predominantly centered on the utilization of tail risk measures in the domains of planning, control, and verification within robotic systems. Nonetheless, it is crucial to emphasize the broader applicability and potential impact of these notions. In this section, we detail several emerging areas that have gained substantial attention.

\subsection{Risk-Aware Learning}
There exists a rich body of literature exploring the integration of tail risk measures within learning paradigms such as reinforcement learning, supervised learning, and unsupervised learning. These studies delve into diverse topics ranging from risk-sensitive reward functions and policy optimization to risk-aware feature learning and model training. Such research underscores the versatile role of tail risk measures in not only guiding robotic behavior in uncertain environments but also enabling robots to learn and adapt in a risk-aware manner over time. Thus, to provide a comprehensive overview of the role of tail risk measures in robotics, it is crucial to shed light on their applications in learning-based contexts as well.

The recent exploration into risk-averse reinforcement learning is well encapsulated by the work of Greenberg et al.~\cite{greenberg2022efficient}. They emphasized the challenges of optimizing risk measures, as conventional methods often overlook high-return strategies. To address this, they proposed a soft risk mechanism coupled with a Cross-Entropy module for efficient risk sampling. This innovative method, while maintaining risk aversion, demonstrated improved risk aversion across diverse benchmarks, setting a precedent for future exploration in this realm.  In another interesting direction, Lacotte et al.~\cite{lacotte2019risk} delved into a risk-sensitive Generative Adversarial Imitation Learning (GAIL) approach aimed to perform as well as or better than the expert regarding a risk profile. 

Focusing on risk-constrained reinforcement learning, Chow et al.~\cite{chow2017risk} developed algorithms for risk-constrained MDPs, using chance constraints or CVaR as the risk representation. Their work represents an important step towards understanding and implementing risk constraints in RL and how these can be used for practical applications.  Finally, Kose and Ruszczynski's work~\cite{kose2021risk} proposed a novel reinforcement learning methodology employing a Markov coherent dynamic risk measure. This work provided new risk-averse counterparts for basic and multistep methods of temporal differences, paving the way for future exploration in risk-averse learning methodologies.

Despite the notable strides made in these studies, \addendum{there are still many open problems in risk-aware reinforcement learning.  One that specifically ties into the next section includes risk-aware reinforcement learning when the underlying distribution changes over time.  Such nonstationarity would frustrate, for example, immediate applications of distributional reinforcement learning for policy development, as such approaches aim to learn the underlying distribution to inform future risk-aware action choices.  However, we do expect such nonstationarity in real-world contexts, \textit{e.g.} changing wind patterns over time for a drone.  Keeping with the realistic theme, most risk-aware reinforcement learning approaches are sample complex.  While this is fine in a simulation or lab-based setting, how do we efficiently learn such policies in real-world contexts where experiments are not cheap?  Alternatively, how might we transition the learned simulation or lab-based policies to real-world contexts where disturbance distributions are almost certainly different?}

\subsection{Risk-Awareness with Nonstationary and Dependent Data}
This second area has garnered substantial interest insofar as it breaks the assumptions in the works discussed in this survey.  Namely, the vast majority of the work discussed has all centered around risk-aware methodologies where either 1) the underlying distribution was known either exactly or in a distributionally-robust sense, or 2) the distribution is queryable in a sample-based fashion, where the algorithm receives independent samples.  What happens when either of these assumptions fails to hold?  This is the central question in a new area of work that is just beginning in the risk-aware space, and for important reasons as well.  Consider a robot ambulating over uneven terrain.  As the robot traverses the space, any recording of the unevenness of the terrain would correspond to samples from a nonstationary distribution, and if the robot's controller builds a map of the terrain with this information and chooses actions predicated on this map, then successive data is necessarily not independent.

When the underlying distribution of the uncertainty changes during a motion planning task, we would ideally like to understand the level of this shift, so that we can account for it in our risk-aware planners. There has been a push towards identifying out-of-distribution data for learning-based tasks~\cite{sinha2022systemlevel, pmlr-v164-farid22a}. In~\cite{pmlr-v164-farid22a}, the authors study task-driven OOD detection using Probably Approximately Correct (PAC)-Bayes theory for training the robot. The PAC-Bayes procedure provides a performance bound such that violating this bound signals that the robot is operating in an OOD environment. 

In addition to \textit{detection} of OOD scenarios, we would also like to \textit{respond} to such scenarios online. Here, distribution-free prediction schemes like those offered by conformal prediction are gaining more traction~\cite{shafer2008tutorial}.  As these tools offer ways of generating probabilistically accurate predictors provided streams of, potentially non-independent data, these predictors have been used for motion planning~\cite{dixit2023adaptive,lindemann2022safe,luo2022sample,ren2023robots}, confidence regions for learned classifiers~\cite{vazquez2022conformal,papadopoulos2009reliable}, and even for risk-aware decision making, though not in a tail risk sense~\cite{ramakrishna2022risk}. If we can identify and adapt to distribution shifts in a risk-aware manner, we can enable robotic systems to react to data drift, unseen data, or spurious correlations~\cite{wiles2021finegrained}. By dynamically adjusting the risk level to adapt to the changing uncertainty distribution and guaranteeing the desired level of safety for the motion planning task, robots can operate in a wider array of unstructured environments while guaranteeing safety, task completion, and efficiency.

% \newpage

\bibliographystyle{unsrtnat}
\bibliography{bib_works.bib}

% \processdelayedfloats % place endfloats here, before the sidebars

%%%%%%%%%%%%%%%%%%%%%%%%%%%%%%%%%%%%%%%%%%%%%%%%
%%%%%%%%%%%%%%%%%%%%%%%%%%%%%%%%%%%%%%%%%%%%%%%%
%%%%%%%%%%%%%%%%%%%%%%%%%%%%%%%%%%%%%%%%%%%%%%%%
% \sidebars % changes numbering scheme

% \clearpage
% \newpage

% \section[Example Sidebar]{Sidebar: Example Sidebar}
% \label{sb:example_sb}
% Follow all normal labeling and referencing schemes.  Any new references that did not pop up in the main document will be cited in the secondary bibliography, to-be-called for each sidebar.
% \begin{thebibliography}{10}
% 	\bibitem[S1]{S1} Random reference number one generated for a sidebar, 2016
% 	\bibitem[S2]{S2} Random reference number two generated for a sidebar, 2016
% \end{thebibliography}
% \newpage

%%%%%%%%%%%%%%%%%%%%%%%%%%%%%%%%%%%%%%%%%%%%%%%%
%%%%%%%%%%%%%%%%%%%%%%%%%%%%%%%%%%%%%%%%%%%%%%%%
%%%%%%%%%%%%%%%%%%%%%%%%%%%%%%%%%%%%%%%%%%%%%%%%

\section{Author Biographies}
\balance
\begin{IEEEbiography}
{\textbf{Prithvi Akella}} received the B.S. degree in Mechanical Engineering from the University of California, Berkeley, in 2018, and the M.S. and Ph.D. degrees in Mechanical Engineering from California Institute of Technology in 2020 and 2023, respectively.  He is the recipient of the Bell Family Graduate Fellowship in Engineering and Applied Sciences.  His current research focuses on the automated test and evaluation of cyber-physical systems.
\spacing
\end{IEEEbiography}

\begin{IEEEbiography}
{\textbf{Anushri Dixit}} is an Assistant Professor in the Department of Mechanical \& Aerospace Engineering at the University of California, Los Angeles. She received her Ph.D. in Control and Dynamical Systems from California Institute of Technology in 2023 and her B.S. in Electrical Engineering from Georgia Institute of Technology in 2017. She was a postdoctoral researcher in the Department of Mechanical \& Aerospace Engineering at Princeton University from 2023-2024. Her research focuses on motion planning and control of robots in unstructured environments while accounting for uncertainty in a principled manner. She has received the Outstanding Student Paper Award at the Conference on Decision and Control, Best Student Paper Award at the Conference of Robot Learning (as a co-author), and was selected as a Rising Star in Data Science by The University of Chicago.  
\spacing
\end{IEEEbiography}

\begin{IEEEbiography}
{\textbf{Mohamadreza Ahmadi}} is currently a technical lead in planning at Gatik AI, Mountain View, CA. He finished his DPhil (PhD) in Engineering Science in November, 2016 at the University of Oxford, UK, as a Clarendon Scholar. His PhD was followed by research positions at the University of Texas, Austin, TX, the Center for Autonomous Systems and Technologies (CAST) at the California Institute of Technology, Pasadena, CA,  NASA Jet Propulsion Laboratory, La Canada, CA, and TuSimple, La Jolla, CA. He is the recipient of the Sloan-Robinson Engineering Fellowship, an Edgell-Sheppee Award, and an ICES Postdoctoral Fellowship. His current research is on planning and control under uncertainty with application to autonomous driving, in particular, short-haul to long-haul trucks.
\spacing
\end{IEEEbiography}

\begin{IEEEbiography}
{\textbf{Lars Lindemann}} is currently an Assistant Professor in the Thomas Lord Department of Computer Science at the University of Southern California where he is also a member of the Ming Hsieh Department of Electrical and Computer Engineering (by courtesy), the Robotics and Autonomous Systems Center, and the Center for Autonomy and Artificial Intelligence. Between 2020 and 2022, he was a Postdoctoral Fellow in the Department of Electrical and Systems Engineering at the University of Pennsylvania. He received the Ph.D. degree in Electrical Engineering from KTH Royal Institute of Technology in 2020.  His research interests include systems and control theory, formal methods, and autonomous systems. Professor Lindemann received the Outstanding Student Paper Award at the 58th IEEE Conference on Decision and Control and the Student Best Paper Award (as a co-advisor) at the 60th IEEE Conference on Decision and Control. He was finalist for the Best Paper Award at the 2022 Conference on Hybrid Systems: Computation and Control and for the Best Student Paper Award at the 2018 American Control Conference.
\spacing
\end{IEEEbiography}

\begin{IEEEbiography}
{\textbf{Margaret P. Chapman}} is an Assistant Professor with the Edward S. Rogers Sr. Department of Electrical and Computer Engineering at the University of Toronto. Her research focuses on risk-aware control theory, and she also investigates different control-theoretic applications, including leukemia treatment. Margaret earned her Ph.D. degree in Electrical Engineering and Computer Sciences (EECS) from the University of California Berkeley (UC Berkeley) in May 2020. In 2021, Margaret received the Leon O. Chua Award for outstanding achievement in nonlinear science from EECS at UC Berkeley, and in 2023, Margaret received the Connaught New Researcher Award from the Office of the Vice-President, Research and Innovation at the University of Toronto. 
\spacing
\end{IEEEbiography}

\begin{IEEEbiography}
{\textbf{George J. Pappas}} received the Ph.D. degree in electrical engineering and computer sciences from the University of California, Berkeley, Berkeley, CA, USA, in 1998. He is currently the Joseph Moore Professor in and the chair of the Department of Electrical and Systems Engineering, University of Pennsylvania, Philadelphia, PA 19104, USA. He also holds a secondary appointment with the Department of Computer and Information Sciences and the Department of Mechanical Engineering and Applied Mechanics. He is a member of the General Robotics, Automation, Sensing, and Perception Lab and the Penn Research in Embedded Computing and Integrated Systems Engineering Center. He was previously the deputy dean for research with the School of Engineering and Applied Science. His research interests include control theory and, in particular, hybrid systems, embedded systems, cyberphysical systems, and hierarchical and distributed control systems, with applications to unmanned aerial vehicles, distributed robotics, green buildings, and biomolecular networks. He was a recipient of various awards, such as the Antonio Ruberti Young Researcher Prize, the IEEE Control Systems Society George S. Axelby Award, the O. Hugo Schuck Best Paper Award, the George H. Heilmeier Award, the National Science Founda- tion Presidential Early Career Award for Scientists and Engineers, and numerous best student papers awards. He is a Fellow of IEEE. 
\spacing
\end{IEEEbiography}

\begin{IEEEbiography}
{\textbf{Aaron D. Ames}} received the B.S. degree in mechanical engineering and the B.A. degree in mathematics from the University of St. Thomas, Saint Paul, MN, USA in 2001, and the M.A. degree in mathematics and the Ph.D. degree in electrical engineering and computer sciences from the University of California, Berkeley, CA, USA, in 2006. From 2006 to 2008, he served as a PostDoctoral Scholar in control and dynamical systems with the California Institute of Technology (Caltech), Pasadena, CA, USA. In 2008, he began his faculty career at Texas A\&M University, College Station, TX, USA. He was an Associate Professor with the Woodruff School of Mechanical Engineering and the School of Electrical and Computer Engineering, Georgia Institute of Technology, Atlanta, GA, USA. Since 2017, he has been a Bren Professor of Mechanical and Civil Engineering and Control and Dynamical Systems at Caltech. His research interests include the areas of robotics, nonlinear, safety-critical control, and hybrid systems, with a special focus on applications to dynamic robots—both formally and through experimental validation. Dr. Ames was a recipient of the 2005 Leon O. Chua Award for Achievement in Nonlinear Science and the 2006 Bernard Friedman Memorial Prize in Applied Mathematics from the University of California, Berkeley. He received the NSF CAREER award in 2010, the 2015 Donald P. Eckman Award, and the 2019 IEEE CSS Antonio Ruberti Young Researcher Prize.
\spacing
\end{IEEEbiography}

\begin{IEEEbiography}
{\textbf{Joel W. Burdick}}, the Richard L. and Dorothy M. Hayman Professor of Mechanical Engineering and Bioengineering, received his undergraduate degrees in mechanical engineering and chemistry from Duke University and M.S. and Ph.D. degrees in mechanical engineering from Stanford University. He has been with the department of mechanical engineering at the Caltech since May 1988, where he has been the recipient of the NSF Presidential Young Investigator award, the Office of Naval Research Young Investigator award, and the Feynman fellowship. He has also received the ASCIT Award for Excellence in Undergraduate Teaching and the GSA Award for Excellence in Graduate Student Education, and received the Popular Mechanics Breakthrough Award in 2011. In addition to mechanical engineering, he is a core faculty of the control and dynamical systems option, as well as a faculty affiliate in the options of bioengineering (BE) and computational and neural systems (CNS). His research interests lie mainly in the areas of robotics, kinematics, and mechanical systems. 
% Current research interests include sensor based robot motion planning, multi-fingered robotic hand manipulation, robotic mobility, and rehabilitation of spinal cord injuries. 
\spacing
\end{IEEEbiography}

\endarticle
\end{document}